\documentclass[11pt]{article}
\usepackage{bbm}
\usepackage{eqnarray,amsmath,amsfonts,amsthm,mathrsfs}
\usepackage{color}
\usepackage{bm}
\usepackage{amssymb}
\usepackage[dvips]{graphicx}
\usepackage{epsfig}
\usepackage{epsf}
\usepackage{float}
\usepackage{subfigure}
\usepackage{amsfonts,amsmath,amsthm,amssymb,graphicx,float,fancyhdr,multirow,hyperref}
\usepackage{booktabs,longtable,authblk}
\usepackage{mathrsfs,hhline}
\usepackage{algorithm}
\usepackage{algorithmic}

\usepackage{enumerate}

\usepackage{appendix}

\usepackage{fancyhdr}
\usepackage[top=2.5cm, bottom=2.5cm, left=3cm, right=3cm]{geometry}
\usepackage{graphicx}
\newtheorem{theorem}{Theorem}
\newtheorem{assumption}{Assumption}
\newtheorem{corollary}{Corollary}
\newtheorem{definition}{Definition}

\newtheorem{lemma}{Lemma}[section]
\newtheorem{proposition}{Proposition}[section]
\newtheorem{remark}{Remark}[section]
\newtheorem{example}{Example}[section]

\newcommand{\proofend}{\hfill $\square$}

\newcommand{\X}{\mathcal{X}}
\newcommand{\lsum}{\sum\limits}

\newcommand{\mc}[1]{\mathcal{#1}}
\newcommand{\mb}[1]{\mathbb{#1}}

\allowdisplaybreaks[4]

\def\begeqn{\begin{equation}}
\def\endeqn{\end{equation}}
\def\begth{\begin{theorem}}
\def\endth{\end{theorem}}
\def\begprop{\begin{proposition}}
\def\endprop{\end{proposition}}
\def\begcor{\begin{corollary}}
\def\endcor{\end{corollary}}
\def\begdef{\begin{definition}}
\def\enddef{\end{definition}}
\def\beglemm{\begin{lemma}}
\def\endlemm{\end{lemma}}
\def\begexm{\begin{example}}
\def\endexm{\end{example}}
\def\begrem{\begin{remark}}
\def\endrem{\end{remark}}
\def\begassum{\begin{assumption}}
\def\endassum{\end{assumption}}

\def\X{\mathcal{X}}

\title{Generalization Analysis of Online Stochastic Gradient Descent for Overparameterized Two-Layer Neural Networks$^\dag$\footnotetext{\dag~ The work by D. Cao and  Z. C. Guo is partially supported by  National Natural Science Foundation of China
		[Project No. 12271473]. The work by L. Shi is partially supported by the National Natural
		Science Foundation of China (Grants No. 12171039, No. 12061160462) and Shanghai Science and Technology
		Program [Project No. 21JC1400600]. All authors contributed equally to this work and are listed
		alphabetically.  The corresponding author is Zheng-Chu Guo. Email addresses:  caodinghao@zju.edu.cn (D. Cao), guozc@zju.edu.cn (Z. C. Guo),  leishi@fudan.edu.cn (L. Shi).}} 

\author{Dinghao Cao$^1$, Zheng-Chu Guo$^1$ and Lei Shi$^2$  \\
	\small $^1$ School of Mathematical Sciences, Zhejiang University, Hangzhou 310058, P. R. China \\
	\small $^2$ Shanghai Key Laboratory for Contemporary Applied Mathematics,\\
	\small School of Mathematical Sciences, Fudan University, Shanghai 200433, P. R. China\\}

\date{}
\begin{document}

	\maketitle
	
	\begin{abstract}
		We investigate the generalization performance of online stochastic gradient descent (SGD) for overparameterized two-layer neural networks under the Neural Tangent Kernel (NTK) regime. Leveraging the NTK approximation together with the convergence theory of stochastic approximation in reproducing kernel Hilbert spaces (RKHSs), we develop a unified framework that connects the optimization dynamics of neural networks with kernel-based statistical learning. This framework enables us to derive sharp convergence rates for the \emph{generalization error of the last iterate} of online SGD. The obtained rates coincide with the optimal statistical rates for kernel methods under appropriate source assumptions on target function. In contrast to existing NTK analyses, which primarily establish optimization convergence or analyze averaged SGD, our results directly characterize the statistical behavior of the practically implemented last-iterate online SGD algorithm for streaming data. Moreover, our analysis substantially relaxes the required degree of overparameterization by reducing the network-width requirement from exponential to polynomial dependence on the sample size or the number of optimization iterations. These results provide a unified theoretical perspective on stochastic optimization, kernel methods, and statistical learning for overparameterized neural networks, while significantly narrowing the gap between existing theory and practical deep learning.
		
	\end{abstract}

	

	\section{Introduction}

	Deep neural networks have achieved remarkable empirical success in a wide range of applications, including computer vision, natural language processing, scientific computing, and reinforcement learning. Despite the nonconvexity and high dimensionality of the associated optimization problems, these models are routinely trained by relatively simple first-order methods. Among them, stochastic gradient descent (SGD) and its variants remain the predominant algorithms because of their scalability, modest computational cost per iteration, and effectiveness on large data sets. A central problem in modern learning theory is therefore to understand when SGD converges for an overparameterized neural network, how rapidly it converges, and how the optimization dynamics interact with the statistical complexity of the underlying learning problem.
	
	Overparameterization plays an important role in explaining the favorable optimization behavior of neural networks. Although increasing the number of parameters appears, at first sight, to make the optimization problem more difficult, sufficiently wide neural networks often possess a more regular local geometry around random initialization. This observation has led to a substantial body of work establishing global convergence guarantees for gradient-based algorithms under suitable width, initialization, and data assumptions. A particularly influential framework is the Neural Tangent Kernel (NTK), introduced by Jacot et al.~\cite{jacot2018neural}. In the infinite-width limit, and under an appropriate parameterization, the evolution of a neural network trained by gradient descent can be described by a kernel method associated with the NTK. For a sufficiently wide but finite network, this suggests that the nonlinear training dynamics may be approximated by linear stochastic dynamics in the reproducing kernel Hilbert space (RKHS) generated by the limiting NTK.
	
	The NTK perspective has provided mathematically tractable tools for studying the optimization and generalization of overparameterized neural networks. Nevertheless, several gaps remain between the available theory and the algorithms used in practice. First, many analyses concern full-batch gradient descent and assume that the complete training sample is available in advance. By contrast, SGD processes one example or one mini-batch at a time and is naturally suited to large-scale and streaming learning problems. The stochastic fluctuations introduced at every iteration make its dynamics substantially more difficult to control than those of batch gradient descent. Second, much of the statistical theory for stochastic approximation concerns an averaged iterate. Averaging reduces the variance accumulated during optimization and often leads to optimal rates, but it produces an algorithm that is different from the standard last-iterate SGD scheme commonly used to train neural networks. Third, existing NTK analyses may impose extremely large network-width requirements, sometimes growing exponentially with the number of iterations or the sample size. Such requirements weaken the practical interpretation of the theoretical guarantees.
	
	The present paper studies these issues for online SGD applied to an overparameterized two-layer neural network. Our objective is not merely to establish convergence to a global minimum of an empirical objective. Instead, we investigate the population prediction error and derive quantitative learning rates determined jointly by the regularity of the regression function, the stochastic optimization dynamics, and the finite width of the neural network. This statistical perspective distinguishes our work from analyses based primarily on the smallest eigenvalue of a finite-sample NTK Gram matrix.
	
	\paragraph{Optimization of overparameterized neural networks.}
	A substantial literature has established that overparameterization can make neural-network optimization theoretically tractable. For two-layer networks, global convergence results for gradient-based methods have been obtained under various assumptions on the data distribution, network structure, and loss function. Representative results include analyses for the squared loss \cite{ge2018learning}, the hinge loss \cite{brutzkus2018sgd}, and the cross-entropy loss \cite{li2018learning}. Convergence guarantees have also been developed for residual networks \cite{li2017convergence}, deep fully connected and convolutional networks \cite{allen2019convergence}, recurrent neural networks \cite{allen2019convergenceRNN}, and overparameterized deep ReLU networks for classification \cite{zou2020gradient}. These results demonstrate that sufficiently wide neural networks can be optimized globally from random initialization under suitable structural conditions.
	
	The NTK framework provides another important route to global convergence. By showing that the empirical tangent kernel remains close to its initialization during training, the nonlinear network dynamics can be related to those of a linearized model. Arora et al.~\cite{arora2019fine}, for example, provided a fine-grained analysis of optimization and generalization for overparameterized two-layer ReLU networks. A common ingredient in this line of work is a positive lower bound on the smallest eigenvalue of the empirical NTK Gram matrix. Such a bound yields rapid optimization of the training error, but its dependence on the sample and the data geometry may become unfavorable as the number of observations increases. Moreover, convergence of the empirical risk alone does not directly characterize the population learning rate.
	
	Recent work has extended the NTK analysis to stochastic and streaming settings. In particular, Xu and Zhu~\cite{xu2024overparametrized} established uniform concentration results for the NTK of multilayer neural networks and proved convergence in expectation for SGD when the observations arrive sequentially. Their analysis incorporates the contributions of multiple hidden layers and provides an important convergence theory for streaming neural-network training. The focus of the present work is complementary: we derive explicit last-iterate population-error rates for two-layer networks under source assumptions, and quantify the width required to transfer the corresponding RKHS rates to the finite neural network.
	
	\paragraph{Statistical learning under the NTK regime.}
	To derive learning rates that remain meaningful as the sample size grows, it is natural to replace finite-sample spectral assumptions by conditions on the integral operator generated by the limiting kernel. This point of view is standard in kernel-based learning and inverse problems. The regularity of the target function can be expressed through a source condition, while the effective dimension or eigenvalue decay of the integral operator associated with kernel characterizes the capacity of the hypothesis space. 
	Together, these assumptions determine the attainable bias--variance tradeoff and the optimal statistical rate.
	
	The convergence of online gradient methods in RKHSs has been extensively studied; see, for example, \cite{ying2008online, yao2010complexity, dieuleveut2016nonparametric, guo2019fast} and the references therein. These studies show that properly chosen step-size schedules and stopping times can provide optimal or nearly optimal convergence rates. However, these results apply to fixed kernel methods, whereas the tangent features of a finite neural network evolve during training. Transferring an RKHS convergence result to a neural network therefore requires uniform control of the discrepancy between the nonlinear network trajectory and its kernel counterpart over the entire stochastic optimization path.
	
	Nitanda and Suzuki~\cite{nitanda2020optimal} made an important step in this direction by studying averaged SGD for overparameterized two-layer neural networks in the NTK regime. Under a source condition with regularity parameter $r$ and a capacity condition with parameter $\beta$, they established fast statistical convergence rates for the averaged iterate. Their analysis does not require a uniform positive lower bound on the empirical NTK Gram matrix, but instead exploits the spectral properties of the limiting NTK operator. However, averaging is essential to their analysis, and the required network width grows exponentially with the  sample size (or equivalently, the total number of iterations). Moreover, their regularization parameter $\lambda$ is chosen as a function of the sample size, which must therefore be specified in advance. Such a design is not applicable in streaming-data settings, where the sample size is unknown a priori. In contrast, our algorithm employs a time-varying regularization parameter that is updated online, making it naturally suitable for streaming data. Furthermore, the last iterate considered in the present paper does not benefit from the variance reduction induced by averaging, requiring a fundamentally different analytical approach.
	
	The role of network width has also received increasing attention. Nguyen and M$\ddot{u}$cke~\cite{nguyen2024many} developed a refined analysis for shallow networks trained by batch gradient descent and obtained polynomial, rather than exponential, width requirements. Their result demonstrates that the very large widths appearing in earlier NTK analyses are not always intrinsic. Our work establishes an analogous improvement for online stochastic optimization: despite the additional randomness of SGD and the absence of iterate averaging, polynomially many neurons suffice to preserve the desired statistical convergence rate.
	
	\paragraph{Finite-width and feature-learning perspectives.}
	The classical NTK approximation describes a lazy-training regime in which the tangent features vary only moderately around initialization. It does not, by itself, explain all forms of representation learning observed in practical neural networks. Recent theoretical studies have therefore investigated regimes in which the learned features evolve nontrivially. For example, Moniri et al.~\cite{moniri2023theory} characterized nonlinear feature learning produced by an initial gradient step in a two-layer network, while more recent multiscale theories describe transitions between kernel-like behavior and data-dependent feature adaptation \cite{rubin2025kernels}. Finite-width proportional-regime analyses have also begun to characterize training and generalization dynamics beyond conventional infinite-width approximations \cite{han2025precise}.
	
	These developments clarify both the usefulness and the limitations of the NTK framework. Feature learning beyond the lazy regime is essential for a complete theory of neural networks, but the NTK regime remains a fundamental and nontrivial benchmark in which one can rigorously connect nonlinear optimization, stochastic approximation, and statistical learning. The objective of this paper is to sharpen this connection for online SGD by obtaining last-iterate rates under polynomial overparameterization. Thus, our results complement, rather than attempt to replace, recent theories of feature learning and finite-width dynamics.
	
	\paragraph{Main contributions.}
	Our main contributions are summarized as follows.
	
	\begin{enumerate}
		\item \textbf{Last-iterate learning rates for online SGD.}
		We establish population-error bounds for the last iterate of online SGD applied to an overparameterized two-layer neural network. Unlike analyses based on Polyak--Ruppert or tail averaging, our results directly concern the non-averaged iterate generated by the standard online algorithm. We obtain capacity-independent optimal convergence rates of the form
			$
			O\left(T^{-\frac{2r}{2r+1}}\right).
			$
		
		\item \textbf{An RKHS-based analysis of nonlinear stochastic dynamics.}
		We combine the NTK approximation with spectral techniques for stochastic gradient methods in RKHSs. 
		This approach avoids relying on a sample-independent positive lower bound for the smallest eigenvalue of the empirical NTK Gram matrix and yields rates that explicitly reflect the statistical structure of the learning problem.
		
		\item \textbf{Polynomial network-width requirements.}
		We substantially relax the overparameterization required to transfer the kernel learning rates to the neural network. Whereas earlier analyses of stochastic training in the NTK regime may require the number of neurons to depend exponentially on the sample size or the number of iterations, our bounds require only polynomial dependence. The improvement follows from refined control of the finite-width approximation error along the stochastic trajectory.
		
		\item \textbf{Handling streaming data.} Our analysis applies to a genuinely online SGD algorithm that naturally accommodates streaming data. Unlike batch gradient descent~\cite{nguyen2024many}, multi-pass SGD, or the averaged SGD studied in~\cite{nitanda2020optimal}, our method does not require the total sample size or the learning horizon to be known in advance. Instead, it employs a time-varying regularization parameter that is updated online, allowing the algorithm to process continuously arriving data while maintaining the established statistical guarantees.
		
	\end{enumerate}
	

	
	The remainder of the paper is organized as follows. Section~\ref{section: preliminary} introduces the two-layer neural network, the online SGD algorithm, and the associated NTK and RKHS. Section~\ref{section: main results} states the main convergence results and discusses their consequences, including the dependence of the rates on the regularity, and width parameters. Section~\ref{section: proof of main results} presents the proofs of the main theorems. Auxiliary estimates and technical proofs are collected in Section~\ref{section: appendix}.

\section{Preliminary}\label{section: preliminary}
In this paper, we consider the supervised regression problem, whose goal is to learn a functional relationship from random samples for predicting future observations. Let $\X\subset\mathbb{R}^d$ be a compact metric space, let $\mathcal{Y}\subseteq\mathbb{R}$, and let $\rho$ be a Borel probability measure on $\mathcal{Z}=\X\times\mathcal{Y}$. For each pair $(x,y)\in\mathcal{Z}$, the variable $x$ denotes the input and $y$ the corresponding output. Given an input $x$, we seek a measurable function $g:\X\rightarrow\mathcal{Y}$ that accurately predicts the output $y$.

Throughout this paper, prediction accuracy is measured by the squared loss
\(
(g(x)-y)^2.
\)
Accordingly, the performance of a predictor $g$ is quantified by its \emph{expected risk}
\[
\mathcal{E}(g)
=
\int_{\X\times\mathcal{Y}}
(g(x)-y)^2\,d\rho(x,y).
\]
It is well known that the minimizer of $\mathcal{E}(g)$ over all measurable functions is the \emph{regression function}
\begin{equation}\label{eq:regression_function}
	g_{\rho}(x)
	=
	\int_{\mathcal{Y}} y\,d\rho(y\mid x),
	\qquad x\in\X,
\end{equation}
where $\rho(\cdot\mid x)$ denotes the conditional distribution of $y$ given $x$ induced by $\rho$.

In statistical learning, the underlying distribution $\rho$ is unknown, and one only has access to independently drawn samples from $\rho$. The objective is therefore to construct an estimator of the regression function $g_{\rho}$ from the observed data. Regression has been extensively studied in statistical learning theory and machine learning; see, for example, \cite{cucker2007learning,steinwart2008support} and the references therein.

The quality of an estimator $g$ is typically evaluated by its \emph{excess risk}, defined as
\[
\mathcal{E}(g)-\mathcal{E}(g_{\rho})
=
\|g-g_{\rho}\|_{\rho}^{2},
\]
where $\|\cdot\|_{\rho}$ denotes the norm of the Hilbert space $\mathcal{L}^2_{\rho_{\X}}$ associated with the marginal distribution $\rho_{\X}$ of $\rho$, whose inner product is given by
\(
\langle u,v\rangle_{\rho}
=
\int_{\X}u(x)v(x)\,d\rho_{\X}(x),
u,v\in\mathcal{L}^2_{\rho_{\X}}.
\)
The excess risk measures the statistical error of the estimator relative to the optimal predictor $g_{\rho}$.

In this work, the estimator $g$ is obtained by training an overparameterized two-layer neural network using online SGD. The neural network model and the corresponding learning algorithm are introduced in the following subsection.

\subsection{Two-layer Neural Networks}\label{subsection: two layer neural networks}

Let $M\in\mathbb{N}_+$ denote the width of the network (i.e., the number of hidden neurons). Let
\[
a=(a_1,\ldots,a_M)^\top\in\mathbb{R}^M,\qquad
B=(b_1^\top,\ldots,b_M^\top)^\top\in\mathbb{R}^{dM},
\]
and
\[
c=(c_1,\ldots,c_M)^\top\in\mathbb{R}^M
\]
be the parameters of the output layer, the hidden layer, and the bias terms, respectively, where $a_r\in\mathbb{R}$, $b_r\in\mathbb{R}^d$, and $c_r\in\mathbb{R}$ for $1\le r\le M$. We denote the collection of all trainable parameters by
\[
\Theta=(a,B,c)\in\mathbb{R}^{M(d+2)}.
\]

Throughout this paper, we consider the two-layer neural network
\begin{equation}\label{equation: two layer neural networks}
	g_{\Theta}(x)
	=
	\frac{1}{\sqrt{M}}
	\sum_{r=1}^{M}
	a_r
	\sigma(b_r^\top x+\gamma c_r),
\end{equation}
where $\sigma:\mathbb{R}\rightarrow\mathbb{R}$ is the activation function and $\gamma>0$ is a scaling parameter for the bias terms.

Training the neural network amounts to finding suitable parameters $\Theta$ that minimize the prediction error. In this paper, the parameters are optimized by online SGD, which requires an initial parameter vector. We adopt the \emph{symmetric initialization} proposed in \cite{su2019learning}. Its main advantage is that the initial network output satisfies
\[
g_{\Theta^{(0)}}(x)\equiv0,
\]
which considerably simplifies the theoretical analysis.

Let
\[
a^{(0)}
=
(a_1^{(0)},\ldots,a_M^{(0)})^\top,\quad
B^{(0)}
=
(b_1^{(0)\top},\ldots,b_M^{(0)\top})^\top,\quad
c^{(0)}
=
(c_1^{(0)},\ldots,c_M^{(0)})^\top.
\]
Throughout the paper, we assume that $M$ is even. The output-layer parameters are initialized by
\[
a_r^{(0)}=
\begin{cases}
	R,&1\le r\le M/2,\\
	-R,&M/2<r\le M,
\end{cases}
\]
where $R>0$ is a fixed constant.

Let $\mu_0$ denote the uniform distribution on the unit sphere
\[
\mathbb{S}^{d-1}
=
\{b\in\mathbb{R}^d:\|b\|_2=1\}.
\]
The hidden-layer parameters are initialized according to
\[
b_r^{(0)}
=
b_{r+M/2}^{(0)}
\overset{\mathrm{i.i.d.}}{\sim}
\mu_0,
\qquad
1\le r\le M/2,
\]
while all bias parameters are initialized by
\[
c_r^{(0)}=0,
\qquad
1\le r\le M.
\]

\subsection{Online Stochastic Gradient Descent}

We consider the following regularized expected-risk minimization problem:
\[
\min_{\Theta}
\left\{
\mathcal{E}(g_\Theta)
+
\lambda
\left(
\|a-a^{(0)}\|_2^2
+
\|B-B^{(0)}\|_2^2
+
\|c-c^{(0)}\|_2^2
\right)
\right\},
\]
where $\lambda>0$ is the regularization parameter, and  $\|\cdot\|_2$ denotes the $\ell^2$ norm for vectors in $\mathbb{R}^d$ or $\mathbb{R}^{dM}$. The regularization term penalizes the deviation of the parameters from their initialization, thereby keeping the optimization trajectory in a neighborhood where the neural tangent kernel approximation remains valid; see \cite{su2019learning}.

To solve the above optimization problem, we employ online stochastic gradient descent. Let
\[
\Theta^{(t)}
=
(a^{(t)},B^{(t)},c^{(t)})
\]
denote the parameter vector at iteration $t$, with
\(
\Theta^{(1)}=\Theta^{(0)}.
\)
At each iteration $t\ge1$, a fresh sample
\(
(x_t,y_t)\sim\rho
\)
is independently drawn from the underlying distribution, and the parameters are updated according to
\begin{align*}
	a_r^{(t+1)}-a_r^{(0)}
	&=
	(1-\eta_t\lambda_t)
	(a_r^{(t)}-a_r^{(0)})
	-
	\eta_tM^{-1/2}
	(g_{\Theta^{(t)}}(x_t)-y_t)
	\sigma(b_r^{(t)\top}x_t+\gamma c_r^{(t)}),
	\\[2mm]
	b_r^{(t+1)}-b_r^{(0)}
	&=
	(1-\eta_t\lambda_t)
	(b_r^{(t)}-b_r^{(0)})
	-
	\eta_tM^{-1/2}
	(g_{\Theta^{(t)}}(x_t)-y_t)
	a_r^{(t)}
	\sigma'(b_r^{(t)\top}x_t+\gamma c_r^{(t)})
	x_t,
	\\[2mm]
	c_r^{(t+1)}-c_r^{(0)}
	&=
	(1-\eta_t\lambda_t)
	(c_r^{(t)}-c_r^{(0)})
	-
	\eta_tM^{-1/2}
	(g_{\Theta^{(t)}}(x_t)-y_t)
	a_r^{(t)}
	\gamma
	\sigma'(b_r^{(t)\top}x_t+\gamma c_r^{(t)}),
\end{align*}
where $\eta_t>0$ is the learning rate and $\lambda_t>0$ is the regularization parameter.

Unlike existing analyses that employ a fixed regularization parameter depending on the total sample size or the number of iterations, we adopt a time-varying regularization schedule,
\[
\eta_t=\eta (t+1)^{-\theta},
\qquad
\lambda_t=\frac{1}{\eta}(t+1)^{-(1-\theta)},
\]
where $\eta>0$ is a constant and
\(
\frac12<\theta<1.
\)
Since $\lambda_t$ depends only on the current iteration, the algorithm does not require prior knowledge of the learning horizon and is therefore naturally applicable to streaming-data settings.

Algorithm~\ref{SGD} summarizes the resulting online SGD procedure. 

\begin{algorithm}[H]
	\caption{Online Stochastic Gradient Descent}
	\label{SGD}
	\begin{algorithmic}[1]
		\REQUIRE Number of iterations $T$, width of network $M$, regularization parameter $\{\lambda_t\}_{t\ge 1}$, step sizes $\{ \eta_t \}_{t\ge 1}$, 
		initial parameters: $\Theta^{(1)} = \Theta^{(0)}=(a^{(0)}, B^{(0)}, c^{(0)})$ and bias parameter $\gamma$
		\ENSURE Estimation for $\Theta^{(T+1)} = (a^{(T+1)}, B^{(T+1)}, c^{(T+1)})$
		\FOR {$t = 1$ to $T$}
		\STATE Randomly draw a sample $z_t=(x_t, y_t) \sim \rho$
		\STATE $\Theta^{(t+1)} \leftarrow \Theta^{(t)}-\eta_t \left(g_{\Theta^{(t)}}(x_{t}) - y_{t}\right) \nabla_{\Theta} g_{\Theta}(x_{t}) \Big|_{\Theta = \Theta^{(t)}}-\eta_t\lambda_t \left(\Theta^{(t)}-\Theta^{(0)}\right)$
		\ENDFOR
	\end{algorithmic}
\end{algorithm}
\subsection{Neural Tangent Kernel}

A key ingredient of our analysis is the \emph{neural tangent kernel} (NTK), which provides a bridge between overparameterized neural networks and kernel methods. It has been shown that, when the network width is sufficiently large, the training dynamics of gradient descent (or stochastic gradient descent) on a two-layer neural network can be accurately approximated by kernel gradient descent in the reproducing kernel Hilbert space (RKHS) associated with the NTK; see, e.g., \cite{jacot2018neural,allen2019convergence,arora2019fine}. This connection enables us to study the statistical behavior of online SGD through the spectral properties of the corresponding kernel operator.

For the two-layer neural network \eqref{equation: two layer neural networks} with the symmetric initialization described in the previous subsection, the limiting NTK is given by
\begin{equation}\label{NTK}
	\begin{aligned}
		k_{\infty}(x,x')
		&=
		\mathbb{E}_{b^{(0)}}
		\!\left[
		\sigma(b^{(0)\top}x)
		\sigma(b^{(0)\top}x')
		\right]
		\\
		&\quad
		+
		R^2(x^\top x'+\gamma^2)
		\mathbb{E}_{b^{(0)}}
		\!\left[
		\sigma'(b^{(0)\top}x)
		\sigma'(b^{(0)\top}x')
		\right],
	\end{aligned}
\end{equation}
where $b^{(0)}\sim\mu_0$.

Since practical neural networks have finite width, we also consider the empirical NTK,
\begin{equation}\label{empirical NTK}
	\begin{aligned}
		k_M(x,x')
		=
		\frac{1}{M}
		\sum_{r=1}^{M}
		\sigma(b_r^{(0)\top}x)
		\sigma(b_r^{(0)\top}x')
		+
		\frac{R^2(x^\top x'+\gamma^2)}{M}
		\sum_{r=1}^{M}
		\sigma'(b_r^{(0)\top}x)
		\sigma'(b_r^{(0)\top}x'),
	\end{aligned}
\end{equation}
which converges to $k_\infty$ as $M\rightarrow\infty$.

Both $k_\infty$ and $k_M$ are continuous, symmetric, and positive semidefinite kernels, and hence are Mercer kernels \cite{cucker2002mathematical}. They induce the RKHSs
\[
\mathcal H_\infty
=
\overline{\mathrm{span}}
\{
k_{\infty,x}(\cdot)=k_\infty(x,\cdot):
x\in\X
\},
\]
and
\[
\mathcal H_M
=
\overline{\mathrm{span}}
\{
k_{M,x}(\cdot)=k_M(x,\cdot):
x\in\X
\},
\]
equipped with the inner products satisfying
\[
\langle
k_{\infty,x},
k_{\infty,x'}
\rangle_{\mathcal H_\infty}
=
k_\infty(x,x'),
\]
and
\[
\langle
k_{M,x},
k_{M,x'}
\rangle_{\mathcal H_M}
=
k_M(x,x').
\]

Consequently, the reproducing property holds:
\[
f(x)
=
\langle
f,
k_{\infty,x}
\rangle_{\mathcal H_\infty},
\qquad
f\in\mathcal H_\infty,
\]
and
\[
g(x)
=
\langle
g,
k_{M,x}
\rangle_{\mathcal H_M},
\qquad
g\in\mathcal H_M.
\]

The RKHS $\mathcal H_\infty$ plays a central role in our analysis. In particular, the regularity of the target function is characterized through a source condition in $\mathcal H_\infty$, while the statistical complexity of the learning problem is quantified by the spectral decay of the integral operator associated with $k_\infty$. These spectral properties ultimately determine the generalization performance of the online SGD algorithm.

\section{Main Results and Discussion}\label{section: main results}
In this section, we present the main theoretical results of the paper under the assumptions introduced below. We then compare our results with the existing literature, highlighting the key improvements of our analysis.

Without loss of generality, we omit the bias terms in the hidden layer. Indeed, by introducing the augmented input and parameter vectors
\[
\bar{x}=(x,\gamma)\in\mathbb{R}^{d+1},
\qquad
\bar{b}_r=(b_r,c_r)\in\mathbb{R}^{d+1},
\]
the two-layer neural network \eqref{equation: two layer neural networks} can be rewritten as
\[
g_{\Theta}(x)
=
\frac{1}{\sqrt{M}}
\sum_{r=1}^{M}
a_r
\sigma(b_r^\top x+\gamma c_r)
=
\frac{1}{\sqrt{M}}
\sum_{r=1}^{M}
a_r
\sigma(\bar{b}_r^\top\bar{x}).
\]
Hence, throughout the remainder of the paper we assume, without loss of generality, that the network has no bias parameters. Accordingly, for $t\ge1$, we write
\[
\Theta^{(t)}
=
\bigl(
a_1^{(t)},\ldots,a_M^{(t)},
b_1^{(t)\top},\ldots,b_M^{(t)\top}
\bigr)^\top
\in
\mathbb{R}^{(d+1)M}.
\]

We begin by stating the assumptions used throughout the paper.
\begin{assumption}\label{assumption activation function conditions}
	The activation function $\sigma:\mathbb{R}\rightarrow\mathbb{R}$ is twice continuously differentiable. Moreover, there exists a constant $C_\sigma>0$ such that
	\(
	\|\sigma'\|_\infty\le C_\sigma,
	\|\sigma''\|_\infty\le C_\sigma,
	\)
	and
	\(
	|\sigma(u)|\le1+|u|,
	\forall\,u\in\mathbb{R}.
		\)
\end{assumption}

Assumption~\ref{assumption activation function conditions} requires the activation function to be sufficiently smooth. It is satisfied by many commonly used activations, including the sigmoid function, the hyperbolic tangent function, and smooth approximations of the ReLU activation.

\begin{assumption}\label{assumption uniform bounded}
	Assume that the marginal distribution $\rho_{\X}$ is non-degenerate and supported on
	\(
	\X=\{x\in\mathbb{R}^d:\|x\|_2\le1\},
	\)
	\(
	\mathcal{Y}\subset[-1,1],
		\) 
	\(
	R=1,\) and \(
	\gamma\in[0,1].
	\)
\end{assumption}

Assumption~\ref{assumption uniform bounded} is a standard boundedness condition in statistical learning theory. The normalization constants are chosen for notational simplicity and can be replaced by arbitrary positive constants. Under Assumptions~\ref{assumption activation function conditions} and \ref{assumption uniform bounded}, the empirical NTK satisfies the uniform bound
\begin{equation}\label{uniform bound of kernel function}
	|k_M(x,x')|
	\le
	4+2C_\sigma^2
	=: \kappa^2,
	\qquad
	\forall\,x,x'\in\X.
\end{equation}

The statistical properties of the proposed algorithm are characterized through the spectral properties of the NTK integral operator. Let $\mathcal{H}_\infty$ denote the RKHS associated with the limiting NTK $k_\infty$. Define the integral operator
\(
L_\infty:\mathcal{L}^2_{\rho_\X}\rightarrow\mathcal{L}^2_{\rho_\X}
\)
by
\begin{equation}\label{integral operator Linfty}
	(L_\infty f)(\cdot)
	=
	\int_{\X}
	k_\infty(x,\cdot)f(x)\,
	d\rho_\X(x),
	\qquad
	f\in\mathcal{L}^2_{\rho_\X}.
\end{equation}
Since $k_\infty$ is a Mercer kernel and $\X$ is compact, $L_\infty$ is a positive, self-adjoint, trace-class operator on $\mathcal{L}^2_{\rho_\X}$. Consequently, there exists an orthonormal eigensystem
\(
\{(\mu_i,\phi_i)\}_{i\ge1},
\)
where
\(
\mu_1\ge\mu_2\ge\cdots\ge0
\)
are the eigenvalues of $L_\infty$. For any $r>0$, the fractional power of $L_\infty$ is defined spectrally by
\[
L_\infty^r
\left(
\sum_{i\ge1}c_i\phi_i
\right)
=
\sum_{i\ge1}
c_i\mu_i^r\phi_i.
\]

Our analysis also relies on standard assumption in statistical learning theory: a \emph{source condition}, which characterizes the regularity of the regression function.

\begin{assumption}[Source condition]
	\label{assumption regularity condition}
	There exists $\frac12<r\le $ such that
	\[
	g_\rho\in
	L_\infty^r(\mathcal{L}^2_{\rho_\X}),
	\]
	or equivalently,
	\[
	\|L_\infty^{-r}g_\rho\|_\rho<\infty.
	\]
\end{assumption}

Assumption~\ref{assumption regularity condition} is a classical regularity assumption in learning theory. It requires the regression function to belong to the range space of $L_\infty^r$,
\[
L_\infty^r(\mathcal{L}^2_{\rho_\X})
=
\left\{
f\in\mathcal{L}^2_{\rho_\X}:
\sum_{i\ge1}
\frac{\langle f,\phi_i\rangle_\rho^2}
{\mu_i^{2r}}
<\infty
\right\},
\]
which quantifies the smoothness of $g_\rho$ relative to the RKHS induced by the NTK. Larger values of $r$ correspond to smoother target functions. Moreover, since
\[
L_\infty^{r_1}
(\mathcal{L}^2_{\rho_\X})
\subseteq
L_\infty^{r_2}
(\mathcal{L}^2_{\rho_\X}),
\qquad
r_1\ge r_2,
\]
larger $r$ imposes a stronger regularity requirement.

Since $\rho_\X$ is non-degenerate, Theorem~4.12 of \cite{cucker2007learning} shows that
$L_\infty^{1/2}$ is an isometric isomorphism from the closure of $\mathcal H_\infty$ in $\mathcal L^2_{\rho_\X}$ onto $\mathcal H_\infty$, namely,
\begin{equation}\label{normrelation2}
	\|f\|_\rho
	=
	\|L_\infty^{1/2}f\|_{\mathcal H_\infty},
	\qquad
	f\in\overline{\mathcal H_\infty}.
\end{equation}
Consequently,
\(
L_\infty^{1/2}(\mathcal L^2_{\rho_\X})
=
\mathcal H_\infty,
\)
and the assumption $r>\frac12$ implies that
\(
g_\rho\in\mathcal H_\infty.
\)

Under the above assumptions, and with appropriate choices of the step size $\eta_t$, regularization parameter $\lambda_t$, and network width $M$, we are ready to present the main result of this paper, which establishes generalization bounds for the last iterate of the online SGD algorithm applied to an overparameterized two-layer neural network. The proof is deferred to Section~\ref{section: proof of main results}. Throughout the paper, for $k\in\mathbb{N}_+$, we denote by
\(
\mathbb{E}_{z_1,\ldots,z_k}
\)
the expectation with respect to the random samples $z_1,\ldots,z_k$, and abbreviate it by $\mathbb{E}_{Z^k}$.

\begin{theorem}
	\label{theorem general result}
	Suppose that Assumptions~\ref{assumption activation function conditions},
	\ref{assumption uniform bounded}, and
	\ref{assumption regularity condition} (with
	$\frac12<r\le1$) hold.
	For $t\in\mathbb{N}_+$, let
	\[
	\eta_t=\eta(t+1)^{-\theta},
	\qquad
	\lambda_t=\frac{1}{\eta}(t+1)^{-(1-\theta)},
	\]
	where $\frac12<\theta<1$ and
	\begin{equation}
		\label{chap7:eq:eat-condition}
		0<\eta<
		\min\left\{
		\frac{1-\theta}{4\kappa^2},
		\sqrt{\frac{2\theta-1}{60\kappa^4\theta}}
		\right\}.
	\end{equation}
	Define
	\[
	M_0(T)
	=
	\left\lceil
	\max\left\{
	T^{6-5\theta},
	\,
	T^{2(2-\theta)},
	\,
	T^{2+2(r+1)(1-\theta)}
	\right\}
	\right\rceil .
	\]
	Then, for every $T\in\mathbb{N}_+$ and every
	$M\ge M_0(T)$, with probability at least
	$1-\delta$,
	\[
	\mathbb{E}_{Z^T}
	\left[
	\left\|
	g_{\Theta^{(T+1)}}-g_{\rho}
	\right\|_{\rho}^{2}
	\right]
	\le
	\widetilde{C}
	\left(
	T^{-2r(1-\theta)}
	+
	T^{-\theta}
	\right)
	\log\frac{2}{\delta},
	\]
where $\widetilde{C}$ is a constant independent of $T$ and $M$, whose explicit expression will be given in the proof.	
\end{theorem}
\begin{remark}
	The probability statement in the above result is with respect to the
	random initialization $\Theta^{(0)}$ of the neural network (and hence the
	randomness induced by the initialization-dependent NTK approximation).
\end{remark}
Choosing
$
\theta=\frac{2r}{2r+1},
$
balances the two dominant error terms
\(
T^{-2r(1-\theta)}
\)
and
\(
T^{-\theta}.
\)
Moreover, a direct calculation shows that
\[
M_0(T)
=
\left\lceil
T^{3+\frac{1}{2r+1}}
\right\rceil.
\]
Consequently, Theorem~\ref{theorem general result} yields the following capacity-independent optimal convergence rate.

\begin{corollary}
	\label{corollary convergence rates}
	Suppose that the assumptions of Theorem~\ref{theorem general result}
	hold, and let
	\[
	\theta=\frac{2r}{2r+1}.
	\]
	If
	\[
	M\geq
	M_0(T)
	=
	\left\lceil
	T^{3+\frac{1}{2r+1}}
	\right\rceil,
	\]
	then, with probability at least $1-\delta$,
	\[
	\mathbb{E}_{Z^T}
	\left[
	\left\|
	g_{\Theta^{(T+1)}}-g_{\rho}
	\right\|_{\rho}^{2}
	\right]
	\le 2\widetilde{C}
	T^{-\frac{2r}{2r+1}}
	\log\frac{2}{\delta},
	\]
	where the constant $\widetilde{C}$ is independent of $T$ and $M$.
\end{corollary}

\noindent \textbf{Remark.} The convergence rate of $O\left(T^{-\frac{2r}{2r + 1}}\right)$ is known to be the capacity-independent minimax optimal rate in statistical learning theory \cite{caponnetto2007optimal}.

Our results are most closely related to those of Nitanda and Suzuki
\cite{nitanda2020optimal}. Under the source condition on the regression function and the capacity condition
\[
b i^{-\beta}
\le
\mu_i(L_\infty)
\le
c i^{-\beta},
\qquad
\beta\ge1,
\]
they established the convergence rate
$
O\!\left(
T^{-\frac{2r\beta}{2r\beta+1}}
\right).
$
The special case $\beta=1$ corresponds to the capacity-independent setting, yielding the minimax optimal convergence rate
$
O\!\left(T^{-\frac{2r}{2r+1}}\right).
$
Compared with \cite{nitanda2020optimal}, the present work differs in several important aspects.
\begin{itemize}
\item First, we establish the convergence rate for the \emph{last iterate} of online SGD, whereas \cite{nitanda2020optimal} analyzes the averaged iterate
\[
g_{\bar{\Theta}^{(T+1)}},
\qquad
\bar{\Theta}^{(T+1)}
=
\frac{1}{T+1}
\sum_{t=1}^{T+1}\Theta^{(t)}.
\]
Although iterate averaging is a classical variance-reduction technique and often improves the robustness of stochastic optimization methods
\cite{nemirovski2009robust,rakhlin2011making,yao2010complexity}, empirical evidence suggests that averaging only the last few iterates, or even returning the last iterate alone, can achieve comparable or even better performance in many applications
\cite{rakhlin2011making,shalev2007pegasos}. Our theoretical results provide further support for the effectiveness of the last iterate in training overparameterized neural networks.

\item Second, our analysis only requires the regularity condition on the target function and does not rely on any capacity assumption on the RKHS. Consequently, our result recovers the capacity-independent minimax optimal convergence rate without imposing additional assumptions on the eigenvalue decay of the infinite-width NTK.

\item Third, our algorithm employs both decaying step sizes and decaying regularization parameters,
$
\eta_t=\eta(t+1)^{-\theta},
\lambda_t=\frac{1}{\eta}(t+1)^{-(1-\theta)},
$
which makes it naturally applicable to streaming-data settings, since neither parameter depends on the total number of samples or iterations. In contrast, the regularization parameter in \cite{nitanda2020optimal} depends explicitly on the total number of iterations and therefore requires this quantity to be known in advance.

\item Finally, our analysis substantially relaxes the required network width. While the theoretical guarantees in \cite{nitanda2020optimal} require the network width to grow exponentially with the number of iterations, Corollary~\ref{corollary convergence rates} establishes the same capacity-independent minimax optimal convergence rate with only a polynomial width requirement,
$
M
=
O\!\left(
T^{3+\frac{1}{2r+1}}
\right).
$
\end{itemize}
This significantly improves the theoretical scalability of overparameterized neural networks.

The analysis of online SGD in RKHS has been extensively studied
\cite{ying2008online,tarres2014online,lei2017convergence,carratino2018learning,guo2019fast,guo2023capacity,guo2023optimality,chen2022online,wang2024differentially}. Most existing works consider optimization within a single RKHS associated with a fixed kernel. In contrast, our analysis simultaneously involves two RKHSs generated by the finite-width NTK $k_M$ and the infinite-width NTK $k_\infty$, respectively. The interaction between these two kernels introduces additional approximation and dynamical errors that are absent in the classical RKHS setting. Among the most closely related works, Tarrès and Yao \cite{tarres2014online} studied a regularized online learning algorithm in an RKHS with time-dependent step sizes and regularization parameters. Under the source condition
$
f_\rho\in L_K^r(\mathcal{L}^2_{\rho_{\mathcal X}}),
\frac12<r\le1,
$
they established the capacity-independent optimal convergence rate
$
O\left(
T^{-\frac{2r}{2r+1}}
\right)
$
with high probability. Our Corollary~\ref{corollary convergence rates} recovers the same capacity-independent optimal rate in the overparameterized neural network setting. Ying and Pontil \cite{ying2008online} investigated the same online learning algorithm with a fixed regularization parameter and obtained the convergence rate
$
O\left(
T^{-\frac{2r}{2r+1}+\varepsilon}
\right)
$
in expectation for any
$(0<\varepsilon<\frac{2r}{2r+1})$. By employing a decaying regularization parameter together with a refined convergence analysis, our result eliminates this arbitrarily small loss and achieves the optimal capacity-independent rate.

Although this paper focuses on two-layer overparameterized neural networks, the analysis developed here may also be useful for studying SGD in deeper neural architectures. Combined with recent approximation results for deep neural networks
\cite{yarotsky2017error,zhou2020universality,lin2022universal},
our techniques may provide a theoretical framework for analyzing the generalization performance of SGD in deep learning. Another promising direction is to extend the present analysis to operator learning based on kernel methods \cite{shi2024learning}.

\section{Proof of Main Results}\label{section: proof of main results}

In this section, we present the proofs of our main results, namely Theorem~\ref{theorem general result} and Corollary~\ref{corollary convergence rates}.

\subsection{Error Decomposition}\label{subsection: error decomposition}

Our objective is to estimate the expected generalization error
\[
\mathbb{E}_{Z^T}\!\left[\left\| g_{\Theta^{(T+1)}}-g_{\rho}\right\|_{\rho}^{2}\right].
\]
To facilitate the analysis, we first decompose this quantity into several components, each of which can be estimated separately. To this end, we introduce a number of fundamental tools and concepts from the convergence analysis of learning algorithms in reproducing kernel Hilbert spaces (RKHSs). These results will serve as the basis of our analysis and will be used repeatedly throughout the remainder of the proofs.

Let $M\in\mathbb{N}_{+}$, and denote by
\[
k_{M,x}=k_M(x,\cdot)
\]
the representer associated with $x$. Analogous to the integral operator $L_{\infty}$ defined in (\ref{integral operator Linfty}), we define the integral operator $L_M:\mathcal{L}^2_{\rho_{\X}}\rightarrow\mathcal{L}^2_{\rho_{\X}}$ associated with the kernel $k_M$ by
\[
L_Mf(\cdot)
=
\int_{\X}f(x)k_M(x,\cdot)\,d\rho_{\X}(x),
\qquad
\forall f\in\mathcal{L}^2_{\rho_{\X}}.
\]

It is well known that, if $\mathrm{Ker}(L_M)=\{0\}$, then the operator
\[
L_M^{1/2}:\mathcal{L}^2_{\rho_{\X}}\rightarrow\mathcal{H}_M
\]
is an isometric isomorphism \cite{cucker2002mathematical}, namely,
\[
\left\langle
L_M^{1/2}f,
L_M^{1/2}g
\right\rangle_{\mathcal{H}_M}
=
\langle
f,g
\rangle_{\mathcal{L}^2_{\rho_{\X}}},
\qquad
\forall f,g\in\mathcal{L}^2_{\rho_{\X}}.
\]

For $\lambda>0$, let $g_{M,\lambda}$ denote the minimizer of the regularized expected risk over $\mathcal{H}_M$, namely,
\begin{equation}\label{equation: definition of g_M,lambda}
	g_{M,\lambda}
	=
	\arg\min_{g\in\mathcal{H}_M}
	\left\{
	\mathcal{E}(g)
	+
	\lambda\|g\|_{\mathcal{H}_M}^{2}
	\right\}.
\end{equation}
Since $\lambda>0$, the minimizer exists uniquely and admits the representation \cite{caponnetto2007optimal}
\begin{align}\label{equation: expression of g_M,lambda}
	g_{M,\lambda}
	=
	(L_M+\lambda I)^{-1}
	\mathbb{E}_{(X,Y)}
	\left[
	Yk_{M,X}
	\right]
	=
	(L_M+\lambda I)^{-1}
	L_Mg_{\rho}.
\end{align}

Similarly, define
\begin{equation}\label{equation: definition of g_infty,lambda}
	g_{\infty,\lambda}
	=
	\arg\min_{g\in\mathcal{H}_{\infty}}
	\left\{
	\mathcal{E}(g)
	+
	\lambda\|g\|_{\mathcal{H}_{\infty}}^{2}
	\right\},
\end{equation}
which satisfies
\begin{align}\label{equation: expression of g_infty,lambda}
	g_{\infty,\lambda}
	=
	(L_{\infty}+\lambda I)^{-1}
	L_{\infty}g_{\rho}.
\end{align}
We next introduce an auxiliary SGD algorithm in the RKHS $\mathcal{H}_M$, commonly referred to as the \emph{reference SGD}. This algorithm serves as an intermediate bridge between the neural network dynamics and kernel gradient descent. Its iterative procedure is summarized in Algorithm~\ref{R-SGD}.


\begin{algorithm}[H]
	\caption{Reference SGD in $\mathcal{H}_M$}
	\label{R-SGD}
	\begin{algorithmic}[1]
		\REQUIRE Number of iterations: $T$, regularization parameter: $\{\lambda_t\}_{t\ge1}$, step sizes: $\{ \eta_t \}_{t\ge1}$
		\ENSURE Estimator $g^{(T+1)}$
		\STATE $g^{(1)} = 0$
		\FOR {$t = 1$ to $T$}
		\STATE Randomly draw a sample $z_t=(x_t, y_t) \sim \rho$
		\STATE $g^{(t+1)} = (1 - \eta_t \lambda_t)g^{(t)} - \eta_t \left(g^{(t)}\left(x_{t}\right)- y_{t}\right) k_{M, x_t}$
		\ENDFOR
	\end{algorithmic}
\end{algorithm}

We are now ready to establish the following error decomposition: 
\begin{equation}\label{equation error decomposition}
	\left\| g_{\Theta^{(T+1)}} - g_{\rho} \right\|_{\rho}^2 \leq 3 \left( \left\| g_{\Theta^{(T+1)}} - g^{(T+1)} \right\|_{\rho}^2 + 
	\left\| g^{(T+1)} - g_{M, \lambda_{T+1}} \right\|_{\rho}^2 + \left\| g_{M, \lambda_{T+1}} - g_{\rho} \right\|_{\rho}^2 \right), 
\end{equation}
The above decomposition admits a natural interpretation.
\begin{itemize}
	\item $\left\| g_{\Theta^{(T+1)}} - g^{(T+1)} \right\|_{\rho}^2$  quantifies the discrepancy between the neural network iterate and the reference SGD iterate. We refer to this term as the {\bf dynamics error}. 
	\item $\left\| g^{(T+1)} - g_{M, \lambda_{T+1}} \right\|_{\rho}^2$ measures the error of the reference SGD relative to the corresponding regularized population solution. We refer to this term as the {\bf reference SGD error}; 
	\item $\| g_{M, \lambda_{T+1}} - g_{\rho} \|_{\rho}^2$ characterizes the error incurred by approximating the target function $g_{\rho}$ with the regularized population solution in the finite-width RKHS $\mathcal{H}_M$. We refer to this term as the {\bf approximation error}. 
\end{itemize}

The remainder of this section is devoted to estimating these three error components separately. Combining the resulting bounds yields the proofs of Theorem~\ref{theorem general result} and Corollary~\ref{corollary convergence rates}. 

\subsection{Estimation of the Dynamical Error}\label{subsection: estimation for dynamics error}
The following proposition provides an estimate of the dynamical error between the
SGD trajectory generated by the finite-width neural network and the reference
SGD trajectory in the RKHS $\mathcal{H}_M$. Specifically, it shows that when the
network width is sufficiently large, the nonlinear SGD dynamics remain close to
their linearized counterpart associated with the NTK.

\begin{proposition}
	\label{proposition: dynamics error}
	Suppose that Assumptions \ref{assumption activation function conditions} and
	\ref{assumption uniform bounded} hold. Let the step size and regularization
	parameter be defined as
	\[
	\eta_t=\eta (t+1)^{-\theta},\qquad
	\lambda_t=\frac{1}{\eta}(t+1)^{-1+\theta},
	\]
	where $\frac12<\theta<1$ and
	$0<\eta<\frac{1-\theta}{4\kappa^2}$.
	If the network width satisfies
	\[
	M\geq M_1(T)=T^{6-5\theta},
	\]
	then
	\begin{align*}
		\left\|g_{\Theta^{(T+1)}}-g^{(T+1)}\right\|_{\rho}^{2}
		\leq
		\left\|g_{\Theta^{(T+1)}}-g^{(T+1)}\right\|_{\infty}^{2}
		\leq
		49C_\sigma^2T^{-\theta}.
	\end{align*}
\end{proposition}

The proof of Proposition \ref{proposition: dynamics error} is presented at the
end of this subsection. To establish this result, we first introduce an
intermediate function obtained by linearizing the neural network around the
initial parameter $\Theta^{(0)}$:
\begin{equation}
	\label{equation:definition of h_{Theta}}
	\begin{aligned}
		h_{\Theta}(x)
		&=
		\left\langle
		\nabla_{\Theta}g_{\Theta}(x)
		\Big|_{\Theta=\Theta^{(0)}},
		\Theta-\Theta^{(0)}
		\right\rangle
		\\
		&=
		\frac1{\sqrt{M}}
		\sum_{r=1}^{M}
		\left(
		(a_r-a_r^{(0)})
		\sigma(b_r^{(0)\top}x)
		+
		a_r^{(0)}
		\sigma'(b_r^{(0)\top}x)
		(b_r-b_r^{(0)})^\top x
		\right).
	\end{aligned}
\end{equation}

Here,
\[
\Theta=(a_1,\ldots,a_M,b_1^\top,\ldots,b_M^\top)^\top
\in\mathbb{R}^{(d+1)M}.
\]
The function $h_\Theta$ represents the first-order Taylor approximation of the
neural network around initialization and plays the role of the corresponding
NTK model. The difference between $g_\Theta$ and $h_\Theta$ quantifies the
nonlinearization error induced by the finite-width network.

Using this linearization, the dynamical error can be decomposed as
\begin{equation}
	\label{error decomposition of dynamics error}
	\begin{aligned}
		\left\|g_{\Theta^{(T+1)}}-g^{(T+1)}\right\|_{\rho}
		&\leq
		\left\|g_{\Theta^{(T+1)}}-g^{(T+1)}\right\|_{\infty}
		\\
		&\leq
		\left\|g_{\Theta^{(T+1)}}-h_{\Theta^{(T+1)}}\right\|_{\infty}
		+
		\left\|h_{\Theta^{(T+1)}}-g^{(T+1)}\right\|_{\infty}.
	\end{aligned}
\end{equation}

where
\[
\|g\|_\infty=\sup_{x\in\mathcal X}|g(x)|.
\]
The first term in (\ref{error decomposition of dynamics error}) corresponds to
the nonlinear approximation error, while the second term measures the deviation
between the linearized neural network dynamics and the reference RKHS-SGD
process.

We next estimate these two terms separately. We first establish a bound on the
network output and the variation of its gradient within a neighborhood of the
initialization.

\begin{lemma}
	\label{lemma: gradient of gtheta}
	Suppose Assumptions \ref{assumption activation function conditions} and
	\ref{assumption uniform bounded} hold. For any $\Theta$ satisfying
	\[
	\|\Theta-\Theta^{(0)}\|_2\leq p
	\]
	for some $p>0$, and $M\geq p^2$, we have
	\begin{equation*}
		\|g_{\Theta}\|_{\infty}
		\leq
		2(1+C_\sigma)p,
	\end{equation*}
	and
	\begin{equation*}
		\sup_{x\in\mathcal X}
		\left\|
		\nabla_{\Theta}g_{\Theta}(x)
		-
		\nabla_{\Theta}g_{\Theta}(x)
		\Big|_{\Theta=\Theta^{(0)}}
		\right\|_2^2
		\leq
		\frac{3C_\sigma^2}{M}p^2 .
	\end{equation*}
\end{lemma}

\begin{proof}
	By the definition of $g_{\Theta}$ and the initialization condition
	$g_{\Theta^{(0)}}(x)=0$, we have
	\[
	g_{\Theta}(x)
	=
	g_{\Theta}(x)-g_{\Theta^{(0)}}(x).
	\]
	Therefore,
	\begin{align*}
		|g_{\Theta}(x)|=
		|g_{\Theta}(x)-g_{\Theta^{(0)}}(x)|&=
		\frac1{\sqrt M}
		\left|
		\sum_{r=1}^{M}
		a_r\sigma(b_r^\top x)
		-
		a_r^{(0)}\sigma(b_r^{(0)\top}x)
		\right|
		\\
		&=
		\frac1{\sqrt M}
		\Bigg|
		\sum_{r=1}^{M}
		\Big[
		(a_r-a_r^{(0)})\sigma(b_r^{(0)\top}x)
		\\
		&\qquad
		+(a_r-a_r^{(0)})
		(\sigma(b_r^\top x)-\sigma(b_r^{(0)\top}x))
		\\
		&\qquad
		+a_r^{(0)}
		(\sigma(b_r^\top x)-\sigma(b_r^{(0)\top}x))
		\Big]
		\Bigg|.
	\end{align*}
	
	Using the boundedness and Lipschitz properties of the activation function in
	Assumption \ref{assumption activation function conditions}, together with
	$\|x\|_2\leq1$, we obtain
	\begin{align*}
		|g_{\Theta}(x)|
		&\leq
		\frac1{\sqrt M}
		\sum_{r=1}^{M}
		\Big[
		|a_r-a_r^{(0)}|
		(1+\|b_r^{(0)}\|_2\|x\|_2)
		\\
		&\qquad
		+
		C_\sigma |a_r-a_r^{(0)}|
		\|b_r-b_r^{(0)}\|_2\|x\|_2
		\\
		&\qquad
		+
		C_\sigma |a_r^{(0)}|
		\|b_r-b_r^{(0)}\|_2\|x\|_2
		\Big].
	\end{align*}
	
	Applying the Cauchy--Schwarz inequality and the boundedness assumption on the
	initial parameters gives
	\begin{align*}
		|g_{\Theta}(x)|
		&\leq
		2
		\left(
		\sum_{r=1}^{M}|a_r-a_r^{(0)}|^2
		\right)^{1/2}+
		\frac{C_\sigma}{\sqrt M}
		\left(
		\sum_{r=1}^{M}|a_r-a_r^{(0)}|^2
		\right)^{1/2}
		\left(
		\sum_{r=1}^{M}\|b_r-b_r^{(0)}\|_2^2
		\right)^{1/2}
		\\
		&\quad+
		C_\sigma
		\left(
		\sum_{r=1}^{M}\|b_r-b_r^{(0)}\|_2^2
		\right)^{1/2}
		\\
		&\leq
		2\|\Theta-\Theta^{(0)}\|_2
		+
		\frac{C_\sigma}{\sqrt M}
		\|\Theta-\Theta^{(0)}\|_2^2
		+
		C_\sigma\|\Theta-\Theta^{(0)}\|_2 .
	\end{align*}
	
	Since $\|\Theta-\Theta^{(0)}\|_2\leq p$ and $M\geq p^2$,
	\[
	\frac{1}{\sqrt M}\|\Theta-\Theta^{(0)}\|_2^2
	\leq p,
	\]
	which implies
	\[
	\|g_{\Theta}\|_{\infty}
	\leq
	2(1+C_\sigma)p .
	\]
	
	It remains to estimate the gradient deviation. By direct calculation,
	\begin{align*}
		&
		\left\|
		\nabla_{\Theta}g_{\Theta}(x)
		-
		\nabla_{\Theta}g_{\Theta}(x)
		\Big|_{\Theta=\Theta^{(0)}}
		\right\|_2^2
		\\
		&=
		\frac1M
		\sum_{r=1}^{M}
		\Big[
		|\sigma(b_r^\top x)-\sigma(b_r^{(0)\top}x)|^2
		+
		|a_r\sigma'(b_r^\top x)
		-
		a_r^{(0)}\sigma'(b_r^{(0)\top}x)|^2
		\|x\|_2^2
		\Big].
	\end{align*}
	
	Using the Lipschitz continuity of $\sigma$ and $\sigma'$,
	\begin{align*}
		&
		\left\|
		\nabla_{\Theta}g_{\Theta}(x)
		-
		\nabla_{\Theta}g_{\Theta}(x)
		\Big|_{\Theta=\Theta^{(0)}}
		\right\|_2^2
		\\
		&\leq
		\frac1M
		\sum_{r=1}^{M}
		\Big[
		C_\sigma^2
		\|b_r-b_r^{(0)}\|_2^2
		+
		2C_\sigma^2|a_r-a_r^{(0)}|^2
		+
		2C_\sigma^2
		\|b_r-b_r^{(0)}\|_2^2
		\Big]
		\\
		&\leq
		\frac{3C_\sigma^2}{M}
		\|\Theta-\Theta^{(0)}\|_2^2 .
	\end{align*}
	
	Therefore,
	\[
	\sup_{x\in\mathcal X}
	\left\|
	\nabla_{\Theta}g_{\Theta}(x)
	-
	\nabla_{\Theta}g_{\Theta}(x)
	\Big|_{\Theta=\Theta^{(0)}}
	\right\|_2^2
	\leq
	\frac{3C_\sigma^2}{M}p^2 .
	\]
	
	This completes the proof.
\end{proof}

\begin{proposition}
	\label{proposition: gtheta-htheta}
	Suppose Assumptions \ref{assumption activation function conditions} and \ref{assumption uniform bounded} hold, for any $\Theta$,
	we have
	\begin{equation*}
		\left\| g_{\Theta} - h_{\Theta} \right\|_{\infty} \leq \frac{4C_\sigma}{\sqrt{M}}	\left\|\Theta - \Theta^{(0)} \right\|_2^2.
	\end{equation*}
\end{proposition}

\begin{proof}
	Recall that
	\[
	g_{\Theta^{(0)}}(x)
	=
	\frac1{\sqrt M}
	\sum_{r=1}^{M}
	a_r^{(0)}
	\sigma(b_r^{(0)\top}x)
	=
	0 .
	\]
	
	Using the definitions of $g_\Theta$ and $h_\Theta$, we have
	\begin{align*}
		g_{\Theta}(x)-h_{\Theta}(x)
		&=
		g_{\Theta}(x)-g_{\Theta^{(0)}}(x)-h_{\Theta}(x)
		\\
		&=
		\frac1{\sqrt M}
		\sum_{r=1}^{M}
		\Big[
		(a_r-a_r^{(0)})
		(\sigma(b_r^\top x)-\sigma(b_r^{(0)\top}x))
		\\
		&\quad+
		a_r^{(0)}
		\big(
		\sigma(b_r^\top x)
		-\sigma(b_r^{(0)\top}x)
		-\sigma'(b_r^{(0)\top}x)
		(b_r-b_r^{(0)})^\top x
		\big)
		\Big].
	\end{align*}
	
	The first term is bounded by the Lipschitz property of $\sigma$:
	\[
	|(a_r-a_r^{(0)})
	(\sigma(b_r^\top x)-\sigma(b_r^{(0)\top}x))|
	\leq
	C_\sigma
	|a_r-a_r^{(0)}|
	\|b_r-b_r^{(0)}\|_2 .
	\]
	
	For the second term, the smoothness of $\sigma$ implies the second-order
	Taylor remainder estimate
	\[
	\left|
	\sigma(u)-\sigma(v)-\sigma'(v)(u-v)
	\right|
	\leq
	C_\sigma |u-v|^2 .
	\]
	
	Consequently,
	\begin{align*}
		|g_{\Theta}(x)-h_{\Theta}(x)|
		&\leq
		\frac{2C_\sigma}{\sqrt M}
		\sum_{r=1}^{M}
		\Big(
		|a_r-a_r^{(0)}|
		\|b_r-b_r^{(0)}\|_2
		+
		\|b_r-b_r^{(0)}\|_2^2
		\Big)
		\\
		&\leq
		\frac{2C_\sigma}{\sqrt M}
		\|\Theta-\Theta^{(0)}\|_2^2
		+
		\frac{2C_\sigma}{\sqrt M}
		\|\Theta-\Theta^{(0)}\|_2^2
		\\
		&\leq
		\frac{4C_\sigma}{\sqrt M}
		\|\Theta-\Theta^{(0)}\|_2^2 .
	\end{align*}
	
	Taking the supremum over $x\in\mathcal X$ yields
	\[
	\|g_{\Theta}-h_{\Theta}\|_\infty
	\leq
	\frac{4C_\sigma}{\sqrt M}
	\|\Theta-\Theta^{(0)}\|_2^2 .
	\]
	
	The proof is complete.
\end{proof}

\begin{proposition}
	\label{prop: bound of the sequence of Theta^t}
	Let $\{\Theta^{(t)}\}_{t\geq1}$ be generated by Algorithm~\ref{SGD}.
	Suppose that Assumptions
	\ref{assumption activation function conditions} and
	\ref{assumption uniform bounded} hold.
	Let
	\[
	\eta_i=\eta(i+1)^{-\theta},
	\qquad
	\lambda_i=\frac1\eta(i+1)^{-1+\theta},
	\]
	where $\frac12<\theta<1$ and
	\[
	0<\eta<\frac{1-\theta}{4\kappa^2}.
	\]
	If
	\[
	M\geq T^{4(1-\theta)},
	\]
	then for every $t\in\mathbb N_T$,
	\begin{equation}
		\label{equation: bound of the sequence Theta_t}
		\left\|\Theta^{(t+1)}-\Theta^{(0)}\right\|_2
		\leq t^{1-\theta}.
	\end{equation}
\end{proposition}
\begin{proof} 
	We prove the result by mathematical induction. 
	
	First, for $t=1,$ since $\Theta^{(1)}=\Theta^{(0)}$, $g_{\Theta^{(0)}}(x)=0,$ $\eta_1=\eta 2^{-\theta},$  and $\lambda_1=\frac{1}{\eta} 2^{-1+\theta},$ 	the online SGD update gives
	\begin{equation*}
		\begin{aligned}
			\left\|\Theta^{(2)} - \Theta^{(0)}\right\|_2&=\left\|(1 - \eta_1 \lambda_1) (\Theta^{(1)} - \Theta^{(0)}) - \eta_1 (g_{\Theta^{(1)}}(x_1) - y_1) \nabla_{\Theta} g_{\Theta}(x_1)\Big|_{\Theta=\Theta^{(1)}} \right\|_2\\
			&=\left\|0 - \eta_1 (0 - y_1) \nabla_{\Theta} g_{\Theta}(x_1)\Big|_{\Theta=\Theta^{(1)}} \right\|_2\\
			&=\eta 2^{-\theta} |y_1| \left\|\nabla_{\Theta} g_{\Theta}(x_1)\Big|_{\Theta=\Theta^{(1)}}\right\|_2.
		\end{aligned}
	\end{equation*}
	For $\left\|\nabla_{\Theta} g_{\Theta}(x_1)\Big|_{\Theta=\Theta^{(1)}}\right\|_2,$ by Assumptions \ref{assumption activation function conditions} and \ref{assumption uniform bounded}, we have 
	\begin{equation*}
		\begin{aligned}
			\left\|\nabla_{\Theta} g_{\Theta}(x_1)\Big|_{\Theta=\Theta^{(1)}}\right\|_2^2&=\frac{1}{M}\sum_{r=1}^M \left[\left(\sigma(b_r^{(1)^\top}x_1)\right)^2+\left(a_r^{(1)}\sigma^{\prime}\left(b_r^{(1)^\top}x_1\right)\right)^2\|x_1\|_2^2 \right]\\
			&\le \frac{1}{M}\sum_{r=1}^M \left[\left(1+|b_r^{(1)^\top}x_1|\right)^2+C_\sigma^2\right]
			\le 4+ C_\sigma^2\le \kappa^2.
		\end{aligned}
	\end{equation*}
	This, combined with the conditions $0 < \eta  < \frac{1-\theta}{4\kappa^2}$ and $|y_1| \leq 1$, implies that
	\begin{equation*}
		\begin{aligned}
			\left\|\Theta^{(2)} - \Theta^{(0)}\right\|_2= \eta 2^{-\theta} |y_1| \left\|\nabla_{\Theta} g_{\Theta}(x_1)\Big|_{\Theta=\Theta^{(1)}}\right\|_2 \le 2^{-\theta}\eta \kappa< \kappa\frac{1-\theta}{4\kappa^2 }< \frac{1-\theta}{4\kappa} \le 1,
		\end{aligned}
	\end{equation*}
	which proves (\ref{equation: bound of the sequence Theta_t}) for $t=1.$ 
	
	As the induction assumption, suppose (\ref{equation: bound of the sequence Theta_t}) holds for all $2\le t\le k$. In order to advance the induction, it is necessary to make an estimation for $\left\|\Theta^{(k+2)} - \Theta^{(0)}\right\|_2,$ i.e., for $t=k+1.$
	Using the online SGD iteration, we obtain
	\begin{equation*}
		\begin{aligned}
		&	\Theta^{(k+2)} - \Theta^{(0)} = (1 - \eta_{k+1} \lambda_{k+1}) (\Theta^{(k+1)} - \Theta^{(0)}) - \eta_{k+1} (g_{\Theta^{(k+1)}}(x_{k+1}) - y_{k+1}) \nabla_{\Theta} g_{\Theta}(x_{k+1}) \Big|_{\Theta = \Theta^{(k+1)}} \\
			&= (1 - \eta_{k+1} \lambda_{k+1}) (\Theta^{(k+1)} - \Theta^{(0)}) - \eta_{k+1} (g_{\Theta^{(k+1)}}(x_{k+1}) - y_{k+1}) \nabla_{\Theta} g_{\Theta}(x_{k+1}) \Big|_{\Theta = \Theta^{(0)}} \\
			&\quad- \eta_{k+1} (g_{\Theta^{(k+1)}}(x_{k+1}) - y_{k+1}) \left( \nabla_{\Theta} g_{\Theta}(x_{k+1}) \Big|_{\Theta = \Theta^{(k+1)}} - \nabla_{\Theta} g_{\Theta}(x_{k+1}) \Big|_{\Theta = \Theta^{(0)}} \right) \\
			&= (1 - \eta_{k+1} \lambda_{k+1}) (\Theta^{(k+1)} - \Theta^{(0)}) - \eta_{k+1} h_{\Theta^{(k+1)}}(x_{k+1}) \nabla_{\Theta} g_{\Theta}(x_{k+1}) \Big|_{\Theta = \Theta^{(0)}} \\
			&\quad- \eta_{k+1} (g_{\Theta^{(k+1)}}(x_{k+1}) - h_{\Theta^{(k+1)}}(x_{k+1})) \nabla_{\Theta} g_{\Theta}(x_{k+1}) \Big|_{\Theta = \Theta^{(0)}} \\
			&\quad- \eta_{k+1} (g_{\Theta^{(k+1)}}(x_{k+1}) - y_{k+1}) \left( \nabla_{\Theta} g_{\Theta}(x_{k+1}) \Big|_{\Theta = \Theta^{(k+1)}} - \nabla_{\Theta} g_{\Theta}(x_{k+1}) \Big|_{\Theta = \Theta^{(0)}} \right) \\
			&\quad+ \eta_{k+1} y_{k+1} \nabla_{\Theta} g_{\Theta}(x_{k+1}) \Big|_{\Theta = \Theta^{(0)}}\\
			&= \left(I - \eta_{k+1} \left(A_{k+1} + \lambda_{k+1} I\right)\right) (\Theta^{(k+1)} - \Theta^{(0)}) - \eta_{k+1} \alpha_{k+1}^{(1)} - \eta_{k+1} \alpha_{k+1}^{(2)} + \eta_{k+1} \alpha_{k+1}^{(3)} \\
			&= -\lsum_{i=1}^{k+1} \eta_i \prod_{j=i+1}^{k+1} \left(I - \eta_j \left(A_j + \lambda_j I\right)\right) (\alpha_i^{(1)} + \alpha_i^{(2)} - \alpha_i^{(3)}),
		\end{aligned}
	\end{equation*}
	where 
	\begin{align*}
		A_i &= \nabla_{\Theta} g_{\Theta}(x_i) \Big|_{\Theta = \Theta^{(0)}} \left(\nabla_{\Theta} g_{\Theta}(x_i) \Big|_{\Theta = \Theta^{(0)}}\right)^T \in \mb{R}^{[(d+1)M] \times [(d+1)M]}, \\
		\alpha_i^{(1)} &= (g_{\Theta^{(i)}}(x_i) - h_{\Theta^{(i)}}(x_i)) \nabla_{\Theta} g_{\Theta}(x_i) \Big|_{\Theta = \Theta^{(0)}}, \\
		\alpha_i^{(2)} &= (g_{\Theta^{(i)}}(x_i) - y_i) \left( \nabla_{\Theta} g_{\Theta}(x_i) \Big|_{\Theta = \Theta^{(i)}} - \nabla_{\Theta} g_{\Theta}(x_i) \Big|_{\Theta = \Theta^{(0)}} \right), \\
		\alpha_i^{(3)} &=  y_i \nabla_{\Theta} g_{\Theta}(x_i) \Big|_{\Theta = \Theta^{(0)}}. 
	\end{align*}
	Since $0 < \eta < \frac{1-\theta}{4\kappa^2}$, and for $j\ge 1,$ $\eta_j=\eta (j+1)^{-\theta},$ $\lambda_j=\frac{1}{\eta}(j+1)^{-1+\theta}$, and $\|A_j\|_{op}\le \kappa^2$. Then $\eta_j \lambda_j=(j+1)^{-1}$, it follows that $\eta_j \left(A_j + \lambda_j I\right)=\eta_j A_j + \eta_j\lambda_j I \preceq\left( \eta (j+1)^{-\theta} \kappa^2+(j+1)^{-1}\right) I\preceq \left( \eta \kappa^2+2^{-1}\right) I \preceq \left( \frac{1-\theta}{4\kappa^2} \kappa^2+2^{-1}\right) I  \preceq I.$ This allows us to establish the following decomposition.
	\begin{equation}\label{induction equation: bound of Theta}
		\begin{aligned}
			&\left\| \Theta^{(k+2)} - \Theta^{(0)} \right\|_2 = \left\| \lsum_{i=1}^{k+1} \eta_i \prod_{j=i+1}^{k+1} \left(I - \eta_j \left(A_j + \lambda_j I\right)\right) (\alpha_i^{(1)} + \alpha_i^{(2)} - \alpha_i^{(3)}) \right\|_2 \\
			&\leq \left\| \lsum_{i=1}^{k+1} \eta_i \prod_{j=i+1}^{k+1} \left(I - \eta_j \left(A_j + \lambda_j I\right)\right) \alpha_i^{(1)} \right\|_2 + \left\| \lsum_{i=1}^{k+1} \eta_i \prod_{j=i+1}^{k+1} \left(I - \eta_j \left(A_j + \lambda_j I\right)\right) \alpha_i^{(2)} \right\|_2 \\
			&\quad+ \left\| \lsum_{i=1}^{k+1} \eta_i \prod_{j=i+1}^{k+1} \left(I - \eta_j \left(A_j + \lambda_j I\right)\right) \alpha_i^{(3)} \right\|_2 \\
			&\leq \lsum_{i=1}^{k+1} \eta_i \left\| \alpha_i^{(1)} \right\|_2 + \lsum_{i=1}^{k+1} \eta_i \left\| \alpha_i^{(2)} \right\|_2 + \lsum_{i=1}^{k+1} \eta_i \left\| \alpha_i^{(3)} \right\|_2. 
		\end{aligned}
	\end{equation}
	We next bound the three terms separately.
	
	For the first term $\lsum_{i=1}^{k+1} \eta_i \left\| \alpha_i^{(1)} \right\|_2$, by $\sup_{x \in \mc{X}}\Big\|\nabla_{\Theta} g_{\Theta}(x)\Big|_{\Theta=\Theta^{(0)}}\Big\|_2^2 = \sup_{x \in \mc{X}} k_M(x, x) \leq \kappa^2$, Proposition \ref{proposition: gtheta-htheta} and the induction assumption, we have 
	\begin{equation*}
		\begin{aligned}
			\lsum_{i=1}^{k+1} \eta_i \left\| \alpha_i^{(1)} \right\|_2 &= \lsum_{i=1}^{k+1} \eta_i \left\| (g_{\Theta^{(i)}}(x_i) - h_{\Theta^{(i)}}(x_i)) \nabla_{\Theta} g_{\Theta}(x_i) \Big|_{\Theta = \Theta^{(0)}} \right\|_2 \\
			&\leq \lsum_{i=1}^{k+1} \eta_i \left\| g_{\Theta^{(i)}} - h_{\Theta^{(i)}} \right\|_{\infty} \left\| \nabla_{\Theta} g_{\Theta}(x_i) \Big|_{\Theta = \Theta^{(0)}} \right\|_2 \\
			&\leq \kappa \lsum_{i=1}^{k+1} \eta_i \frac{4C_{\sigma}}{\sqrt{M}} (i-1)^{2(1-\theta)} \\
			&\leq \frac{4C_{\sigma}\kappa \eta}{\sqrt{M}} \lsum_{i=1}^{k+1} i^{2-3\theta} \\
			&\leq \frac{4C_{\sigma}\kappa \eta}{3(1-\theta)} \frac{(k+1)^{3(1-\theta)}}{\sqrt{M}}. 
		\end{aligned}
	\end{equation*}
	Employing $M \geq T^{4(1-\theta)}$, $\kappa^2 \geq 2C_{\sigma}^2$ and $0 < \eta < \frac{1-\theta}{4C_{\sigma} \kappa}$, there holds 
	\begin{equation*}
		\begin{aligned}
			\lsum_{i=1}^{k+1} \eta_i \left\| \alpha_i^{(1)} \right\|_2 \leq \frac{1}{3} \frac{T^{2(1-\theta)}}{\sqrt{M}} (k+1)^{1-\theta} < \frac{1}{3} (k+1)^{1-\theta}. 
		\end{aligned}
	\end{equation*}
	For the second term $\lsum_{i=1}^{k+1} \eta_i \left\| \alpha_i^{(2)} \right\|_2$, we can use the induction assumption along with Lemma \ref{lemma: gradient of gtheta} and $0\le \eta <\frac{1-\theta}{4\kappa^2}$ to derive
	\begin{equation*}
		\begin{aligned}
			\lsum_{i=1}^{k+1} \eta_i \left\| \alpha_i^{(2)} \right\|_2 &= \lsum_{i=1}^{k+1} \eta_i \left\| (g_{\Theta^{(i)}}(x_i) - y_i) \left( \nabla_{\Theta} g_{\Theta}(x_i) \Big|_{\Theta = \Theta^{(i)}} - \nabla_{\Theta} g_{\Theta}(x_i) \Big|_{\Theta = \Theta^{(0)}} \right) \right\|_2 \\
			&\leq \lsum_{i=1}^{k+1} \eta_i (1 + \left\| g_{\Theta^{(i)}} \right\|_{\infty}) \sup_{x \in \mc{X}} \left\| \nabla_{\Theta} g_{\Theta}(x) \Big|_{\Theta = \Theta^{(i)}} - \nabla_{\Theta} g_{\Theta}(x) \Big|_{\Theta = \Theta^{(0)}} \right\|_2 \\
			&\leq \lsum_{i=1}^{k+1} \eta_i [1 + 2(1+C_{\sigma}) (i-1)^{1-\theta}]  \cdot \frac{\sqrt{3}C_{\sigma}}{\sqrt{M}} (i-1)^{1-\theta} \\
			&\leq \lsum_{i=1}^{k+1} \eta_i [1 + 2(1+C_{\sigma})] i^{1-\theta}  \cdot \frac{\sqrt{3}C_{\sigma}}{\sqrt{M}} i^{1-\theta} \\
			&\leq \frac{\eta \sqrt{3}C_{\sigma}(2C_{\sigma}+3)}{3(1-\theta)} \frac{(k+1)^{3(1-\theta)}}{\sqrt{M}} \\
				&\leq \frac{2 \sqrt{3}\eta\kappa^2}{3(1-\theta)} \frac{(k+1)^{3(1-\theta)}}{\sqrt{M}}\\
			&\leq \frac{1}{3} (k+1)^{1-\theta}, 
		\end{aligned}
	\end{equation*}
	where the last inequality holds due to $M \geq T^{4(1-\theta)}$ and $\frac{2 \sqrt{3}\eta\kappa^2}{1-\theta} <\frac{1-\theta}{4\kappa^2} \cdot\frac{2 \sqrt{3}\kappa^2}{1-\theta}< 1$. 
	
	For the last term $\lsum_{i=1}^{k+1} \eta_i \left\| \alpha_i^{(3)} \right\|_2$, we have
		\begin{align*}
			\lsum_{i=1}^{k+1} \eta_i \left\| \alpha_i^{(3)} \right\|_2 = \lsum_{i=1}^{k+1} \eta_i \left\| y_i \nabla_{\Theta} g_{\Theta}(x_i) \Big|_{\Theta = \Theta^{(0)}} \right\|_2 
			\leq \kappa \lsum_{i=1}^{k+1} \eta_i 
			\leq \frac{\eta \kappa}{1-\theta} (k+1)^{1-\theta} 
			\leq \frac{1}{3} (k+1)^{1-\theta}, 
		\end{align*}
	where the last inequality uses the conditions $0 < \eta < \frac{1-\theta}{4\kappa^2 }$ and $\kappa>1,$ which imply $ \frac{\eta \kappa}{1-\theta}< \frac{1-\theta}{4\kappa^2 }\cdot\frac{ \kappa}{1-\theta}  <\frac{1}{3}$. 
	
	Putting all these three estimates back into (\ref{induction equation: bound of Theta}) yields that 
	\begin{equation*}
		\left\| \Theta^{(k+2)} - \Theta^{(0)} \right\|_2 \leq (k+1)^{1-\theta}. 
	\end{equation*}
	Therefore, by induction, (\ref{equation: bound of the sequence Theta_t}) holds
	for all \(t\in\mathbb N_T\).
\end{proof}

We next estimate the second term
$\left\|h_{\Theta^{(T+1)}}-g^{(T+1)}\right\|_\infty$, in the dynamical error
decomposition \eqref{error decomposition of dynamics error}. 
	\begin{proposition}
	\label{proposition: h_theta-g^T}
	Suppose that Assumptions
	\ref{assumption activation function conditions} and
	\ref{assumption uniform bounded} hold. For any $i\in\mathbb{N}_+$, let
	\[
	\eta_i=\eta(i+1)^{-\theta},
	\qquad
	\lambda_i=\frac{1}{\eta}(i+1)^{-1+\theta},
	\]
	where $\frac{1}{2}<\theta<1$ and
	$
	0<\eta<\frac{1-\theta}{4\kappa^2}.
	$
	If
	\[
	M\geq T^{6-5\theta},
	\]
	then
	\[
	\left\|h_{\Theta^{(T+1)}}-g^{(T+1)}\right\|_\infty
	\leq
	3C_\sigma T^{-\theta/2}.
	\]
\end{proposition}
\begin{proof}
	By the definitions of $h_{\Theta^{(T+1)}}$, $g^{(T+1)}$ and the online SGD update for $\Theta^{(T+1)}$, we obtain 
		\begin{align*}\label{func-recursion-eq}
			&h_{\Theta^{(T+1)}} - g^{(T+1)} =\left\langle \nabla_{\Theta}g_{\Theta}(\cdot)\Big|_{\Theta=\Theta^{(0)}}, \Theta^{(T+1)} - \Theta^{(0)} \right\rangle\\
			&\quad- \left((1 - \eta_T \lambda_T)g^{(T)} - \eta_T \left(g^{(T)}(x_T)-y_{T}\right) k_{M, x_T}\right)\\	
			&=(1 - \eta_T \lambda_T)h_{\Theta^{(T)}} - \eta_T (g_{\Theta^{(T)}}(x_T) - y_T) \left\langle \nabla_{\Theta} g_{\Theta}(\cdot)\Big|_{\Theta=\Theta^{(0)}}, \nabla_{\Theta} g_{\Theta}(x_T) \Big|_{\Theta=\Theta^{(T)}}\right\rangle\\
			&\quad- \left((1 - \eta_T \lambda_T)g^{(T)} - \eta_T \left(g^{(T)}(x_T)-y_{T}\right) k_{M, x_T}\right)\\	
			&=(1 - \eta_T \lambda_T)\left(h_{\Theta^{(T)}} - g^{(T)}\right) - \eta_T (g_{\Theta^{(T)}}(x_T) - y_T) \left\langle \nabla_{\Theta} g_{\Theta}(\cdot)\Big|_{\Theta=\Theta^{(0)}}, \nabla_{\Theta} g_{\Theta}(x_T) \Big|_{\Theta=\Theta^{(T)}}\right\rangle  \\
			&\quad + \eta_T (g^{(T)}(x_T) - y_T) k_{M, x_T}  \\
			&=(1 - \eta_T \lambda_T) \left(h_{\Theta^{(T)}} - g^{(T)}\right) - \eta_T (g_{\Theta^{(T)}}(x_T) - y_T) \left\langle \nabla_{\Theta} g_{\Theta}(\cdot)\Big|_{\Theta=\Theta^{(0)}}, \nabla_{\Theta} g_{\Theta}(x_T) \Big|_{\Theta=\Theta^{(T)}}\right\rangle  \\
			&\quad + \eta_T (g^{(T)}(x_T) - h_{\Theta^{(T)}}(x_T)) k_{M, x_T} + \eta_T (h_{\Theta^{(T)}}(x_T) - y_T) k_{M, x_T}  \\
			&= \left(I - \eta_T(k_{M, x_T} \otimes k_{M, x_T} + \lambda_T I)\right) \left(h_{\Theta^{(T)}} - g^{(T)}\right) \\
			&\quad - \eta_T (g_{\Theta^{(T)}}(x_T) - y_T) \left\langle \nabla_{\Theta} g_{\Theta}(\cdot)\Big|_{\Theta=\Theta^{(0)}}, \nabla_{\Theta} g_{\Theta}(x_T) \Big|_{\Theta=\Theta^{(T)}} - \nabla_{\Theta} g_{\Theta}(x_T) \Big|_{\Theta=\Theta^{(0)}}\right\rangle  \\
			&\quad - \eta_T (g_{\Theta^{(T)}}(x_T) - y_T) \left\langle \nabla_{\Theta} g_{\Theta}(\cdot)\Big|_{\Theta=\Theta^{(0)}}, \nabla_{\Theta} g_{\Theta}(x_T) \Big|_{\Theta=\Theta^{(0)}}\right\rangle  
			+ \eta_T (h_{\Theta^{(T)}}(x_T) - y_T) k_{M, x_T}  \\
			&= \left(I - \eta_T(k_{M, x_T} \otimes k_{M, x_T} + \lambda_T I)\right) \left(h_{\Theta^{(T)}} - g^{(T)}\right) \\
			&\quad - \eta_T (g_{\Theta^{(T)}}(x_T) - y_T) \left\langle \nabla_{\Theta} g_{\Theta}(\cdot)\Big|_{\Theta=\Theta^{(0)}}, \nabla_{\Theta} g_{\Theta}(x_T) \Big|_{\Theta=\Theta^{(T)}} - \nabla_{\Theta} g_{\Theta}(x_T) \Big|_{\Theta=\Theta^{(0)}}\right\rangle  \\
			&\quad - \eta_T (g_{\Theta^{(T)}}(x_T) - h_{\Theta^{(T)}}(x_T)) k_{M, x_T}\\
			&= \left(I - \eta_T(k_{M, x_T} \otimes k_{M, x_T} + \lambda_T I)\right) \left(h_{\Theta^{(T)}} - g^{(T)}\right) 
			- \eta_T \left(\beta_T^{(1)}+\beta_T^{(2)}\right),
		\end{align*}
	where 
	\begin{align*}
		\beta_t^{(1)} &= (g_{\Theta^{(t)}}(x_t) - y_t) \left\langle \nabla_{\Theta} g_{\Theta}(\cdot)\Big|_{\Theta=\Theta^{(0)}}, \nabla_{\Theta} g_{\Theta}(x_t)\Big|_{\Theta=\Theta^{(t)}} - \nabla_{\Theta} g_{\Theta}(x_t)\Big|_{\Theta=\Theta^{(0)}} \right\rangle, \\
		\beta_t^{(2)} &= (g_{\Theta^{(t)}}(x_t) - h_{\Theta^{(t)}}(x_t)) \left\langle \nabla_{\Theta} g_{\Theta}(\cdot)\Big|_{\Theta=\Theta^{(0)}}, \nabla_{\Theta} g_{\Theta}(x_t) \Big|_{\Theta=\Theta^{(0)}}\right\rangle. 
	\end{align*}
	Then we can proceed with the following error decomposition
	\begin{equation}\label{eq: error decompositon of the second term of dynamic error}
	\begin{aligned}
		&\left\| h_{\Theta^{(T+1)}} - g^{(T+1)} \right\|_{\infty} = 
		\left\| \lsum_{t=1}^{T} \eta_t \prod_{i=t+1}^{T} \left(I - \eta_i(k_{M, x_i} \otimes k_{M, x_i} + \lambda_i I)\right) (\beta_t^{(1)} + \beta_t^{(2)}) \right\|_{\infty}\\
		&\le \left\| \lsum_{t=1}^{T} \eta_t \prod_{i=t+1}^{T} \left(I - \eta_i(k_{M, x_i} \otimes k_{M, x_i} + \lambda_i I)\right)\beta_t^{(1)} \right\|_{\infty}+\left\| \lsum_{t=1}^{T} \eta_t \prod_{i=t+1}^{T} \left(I - \eta_i(k_{M, x_i} \otimes k_{M, x_i} + \lambda_i I)\right) \beta_t^{(2)} \right\|_{\infty}\\
		&\le  2\lsum_{t=1}^{T} \eta_t \left\| \beta_t^{(1)} \right\|_{\infty}+ 2\lsum_{t=1}^{T} \eta_t \left\| \beta_t^{(2)} \right\|_{\infty},
	\end{aligned}
	\end{equation}
	The last inequality follows from the following estimate. For $j=1,2$, by the
	specific choices of $\lambda_i=\frac{1}{\eta}(i+1)^{-1+\theta}$ and $\eta_i=\eta (i+1)^{-\theta}$ with $\frac12<\theta<1$ and $0<\eta<\frac{1-\theta}{4\kappa^2}$, we have
	\begin{equation}
		\begin{aligned}
\left\|\left(I - \eta_i(k_{M, x_i} \otimes k_{M, x_i} + \lambda_i I)\right)\beta_t^{(j)}\right\|_\infty&=\left\|\left(I - \eta_i \lambda_i I\right)\beta_t^{(j)}+ \eta_i k_{M, x_i} \otimes k_{M, x_i}\beta_t^{(j)}\right\|_\infty\\
&=\left\|\left(I - \eta_i \lambda_i I\right)\beta_t^{(j)}+ \eta_i \beta_t^{(j)}(x_i) k_{M, x_i}\right\|_\infty\\
&\leq \left(I - \eta_i \lambda_i \right)\left\|\beta_t^{(j)}\right\|_\infty + \eta_i \kappa^2 \left\| \beta_t^{(j)}\right\|_\infty\\
&\le \left\|\beta_t^{(j)}\right\|_\infty + \frac{1-\theta}{4} \left\| \beta_t^{(j)}\right\|_\infty\le 2\left\|\beta_t^{(j)}\right\|_\infty
\end{aligned}
	\end{equation}
		We estimate the two terms on the right-hand side separately. For the first
	term, the Cauchy--Schwarz inequality and the bound
	\[
	\sup_{x\in\mathcal{X}}
	\left\|
	\nabla_\Theta g_\Theta(x)\big|_{\Theta=\Theta^{(0)}}
	\right\|_2
	\leq\kappa
	\]
	imply that
	\begin{align*}
		\left\|\beta_t^{(1)}\right\|_\infty
		&\leq
		\kappa
		\bigl(\left\|g_{\Theta^{(t)}}\right\|_\infty+1\bigr)
		\sup_{x\in\mathcal{X}}
		\left\|
		\nabla_\Theta g_\Theta(x)\big|_{\Theta=\Theta^{(t)}}
		-
		\nabla_\Theta g_\Theta(x)\big|_{\Theta=\Theta^{(0)}}
		\right\|_2.
	\end{align*}
	By Lemma~\ref{lemma: gradient of gtheta} and
	Proposition~\ref{prop: bound of the sequence of Theta^t},
	\[
	\left\|g_{\Theta^{(t)}}\right\|_\infty
	\leq
	2(1+C_\sigma)(t-1)^{1-\theta}
	\]
	and
	\[
	\sup_{x\in\mathcal{X}}
	\left\|
	\nabla_\Theta g_\Theta(x)\big|_{\Theta=\Theta^{(t)}}
	-
	\nabla_\Theta g_\Theta(x)\big|_{\Theta=\Theta^{(0)}}
	\right\|_2
	\leq
	\frac{\sqrt{3}C_\sigma}{\sqrt{M}}(t-1)^{1-\theta}.
	\]
	Consequently,
	\begin{align*}
		\sum_{t=1}^{T}\eta_t\left\|\beta_t^{(1)}\right\|_\infty
		&\leq
		\frac{\sqrt{3}\eta C_\sigma\kappa}{\sqrt{M}}
		\sum_{t=1}^{T}
		t^{-\theta}
		\left[1+2(1+C_\sigma)t^{1-\theta}\right]
		t^{1-\theta}
		\\
		&\leq
		\frac{
			\sqrt{3}\eta C_\sigma\kappa
			\bigl(1+2(1+C_\sigma)\bigr)
		}{\sqrt{M}}
		\sum_{t=1}^{T}t^{2-3\theta}
		\\
		&\leq
		\frac{
			\sqrt{3}\eta C_\sigma\kappa
			\bigl(1+2(1+C_\sigma)\bigr)
		}{3(1-\theta)}
		\frac{T^{3(1-\theta)}}{\sqrt{M}}.
	\end{align*}
	Since $M\geq T^{6-5\theta}$,
	\[
	\frac{T^{3(1-\theta)}}{\sqrt{M}}
	\leq
	T^{-\theta/2}.
	\]
	Moreover, by the definition of $\kappa$,
	\[
	\kappa\bigl(1+2(1+C_\sigma)\bigr)
	\leq
	5\kappa^2.
	\]
	Therefore,
	\begin{equation}
		\sum_{t=1}^{T}\eta_t\left\|\beta_t^{(1)}\right\|_\infty
		\leq
		\frac{5\sqrt{3}\kappa^2\eta C_\sigma}
		{3(1-\theta)}
		T^{-\theta/2}.
		\label{equation: beta1-bound}
	\end{equation}
	
	For the second term, observe that
	\[
	\left\|k_{M,x_t}\right\|_\infty
	=
	\sup_{x\in\mathcal{X}}|k_M(x,x_t)|
	\leq
	\kappa^2.
	\]
	Thus, by Proposition~\ref{proposition: gtheta-htheta} and
	Proposition~\ref{prop: bound of the sequence of Theta^t},
	\begin{align*}
		\sum_{t=1}^{T}\eta_t\left\|\beta_t^{(2)}\right\|_\infty
		&\leq
		\kappa^2
		\sum_{t=1}^{T}
		\eta_t
		\left\|g_{\Theta^{(t)}}-h_{\Theta^{(t)}}\right\|_\infty
		\\
		&\leq
		\frac{4\kappa^2\eta C_\sigma}{\sqrt{M}}
		\sum_{t=1}^{T}t^{2-3\theta}
		\\
		&\leq
		\frac{4\kappa^2\eta C_\sigma}{3(1-\theta)}
		\frac{T^{3(1-\theta)}}{\sqrt{M}}
		\\
		&\leq
		\frac{4\kappa^2\eta C_\sigma}{3(1-\theta)}
		T^{-\theta/2}.
	\end{align*}
	Hence,
	\begin{equation}
		\sum_{t=1}^{T}\eta_t\left\|\beta_t^{(2)}\right\|_\infty
		\leq
		\frac{4\kappa^2\eta C_\sigma}{3(1-\theta)}
		T^{-\theta/2}.
		\label{equation: beta2-bound}
	\end{equation}
	
	Combining \eqref{eq: error decompositon of the second term of dynamic error},
	\eqref{equation: beta1-bound}, and
	\eqref{equation: beta2-bound}, we obtain
	\begin{align*}
		\left\|h_{\Theta^{(T+1)}}-g^{(T+1)}\right\|_\infty
		&\leq
		\frac{2(4+5\sqrt{3})\kappa^2\eta C_\sigma}
		{3(1-\theta)}
		T^{-\theta/2}.
	\end{align*}
	Finally, since
	\[
	\eta<\frac{1-\theta}{4\kappa^2},
	\]
	we have
	\[
	\frac{(8+10\sqrt{3})\kappa^2\eta}
	{3(1-\theta)}
	<
	\frac{8+10\sqrt{3}}{12}
	<3.
	\]
	It follows that
	\[
	\left\|h_{\Theta^{(T+1)}}-g^{(T+1)}\right\|_\infty
	\leq
	3C_\sigma T^{-\theta/2}.
	\]
	This completes the proof.
	\end{proof}
	
	We are now ready to prove
	Proposition~\ref{proposition: dynamics error}.
	
	\begin{proof}[Proof of Proposition~\ref{proposition: dynamics error}]
	By Proposition~\ref{proposition: gtheta-htheta} and
	Proposition~\ref{prop: bound of the sequence of Theta^t}, we have
	\begin{align}
		\left\|
		g_{\Theta^{(T+1)}}-h_{\Theta^{(T+1)}}
		\right\|_\infty
		\leq
		\frac{4C_\sigma}{\sqrt{M}}
		\left\|\Theta^{(T+1)}-\Theta^{(0)}\right\|_2^2
		\leq
		\frac{4C_\sigma}{\sqrt{M}}T^{2(1-\theta)}
		\leq
		4C_\sigma T^{-\theta/2},
		\label{equation:bound of gtheta-htheta}
	\end{align}
	where the last inequality follows whenever
	\[
	M\geq T^{4-3\theta}.
	\]
	
	The parameter-movement estimate in Proposition \ref{prop: bound of the sequence of Theta^t} additionally requires
	\[
	M\geq T^{4(1-\theta)},
	\]
	whereas Proposition~\ref{proposition: h_theta-g^T} requires
	\[
	M\geq T^{6-5\theta}.
	\]
	Since $\frac12<\theta<1$, we have
	\[
	\max\left\{
	T^{4(1-\theta)},
	T^{4-3\theta},
	T^{6-5\theta}
	\right\}
	=
	T^{6-5\theta}.
	\]
	
	Combining \eqref{error decomposition of dynamics error},
	\eqref{equation:bound of gtheta-htheta}, and
	Proposition~\ref{proposition: h_theta-g^T}, we obtain
	\begin{align*}
		\left\|
		g_{\Theta^{(T+1)}}-g^{(T+1)}
		\right\|_\infty
		&\leq
		\left\|
		g_{\Theta^{(T+1)}}-h_{\Theta^{(T+1)}}
		\right\|_\infty
		+
		\left\|
		h_{\Theta^{(T+1)}}-g^{(T+1)}
		\right\|_\infty
		\leq
		7C_\sigma T^{-\theta/2}.
	\end{align*}
	Therefore,
	\[
	\left\|
	g_{\Theta^{(T+1)}}-g^{(T+1)}
	\right\|_\rho^2
	\leq
	\left\|
	g_{\Theta^{(T+1)}}-g^{(T+1)}
	\right\|_\infty^2
	\leq
	49C_\sigma^2T^{-\theta}.
	\]
	This completes the proof.
	\end{proof}

\subsection{Estimation of the Reference SGD Error}\label{subsection: estimation for reference SGD error}
In this subsection, we estimate the reference SGD error 
$\left\|g^{(T+1)}-g_{M, \lambda_{T+1}}\right\|_{\rho}^{2}.$ To this end, we first provide the following error decomposition. 
\begin{lemma}
	\label{chap7:lem:error-decomp-of-convergence-error}
	Let $\left\{ g^{(t)} \right\}$ be generated by Algorithm \ref{R-SGD} and let $g_{M, \lambda_{T+1}}$ be defined by (\ref{equation: expression of g_M,lambda}). Then
	\begin{equation}
		\label{chap7:eq:error-decomp-of-convergence-error}
		\begin{aligned}
			&\quad \mathbb{E}_{Z^{T}} \left[\left\|g^{(T+1)}-g_{M, \lambda_{T+1}}\right\|_{\rho}^{2}\right] \leq 3\left\|\prod_{t=1}^{T}\left(I-\eta_{t}(L_M + \lambda_t I)\right) g_{M, \lambda_1}\right\|_{\rho}^{2} \\
			&+ 3\left\| \sum_{t=1}^T \prod_{i=t+1}^T \left(I - \eta_i(L_M + \lambda_i I)\right) (g_{M, \lambda_{t+1}} - g_{M, \lambda_t}) \right\|_{\rho}^2 \\
			&+ 6\sum_{t=1}^{T} \eta_t^2 \left( 1 + \mathbb{E}_{Z^{t-1}} \left[\left\| g^{(t)} \right\|_{\infty}^2\right] \right) \textnormal{Tr} \left[ L_M^2 \prod_{i=t+1}^{T}\left(I-\eta_{i}\left(L_M+\lambda_i I\right)\right)^2 \right]. 
		\end{aligned}
	\end{equation}
	Here and throughout this subsection, a product over an empty index set is
understood as the identity operator; in particular,
\[
\prod_{i=T+1}^{T}
\left(I-\eta_i(L_M+\lambda_i I)\right)^2
=I.
\]
\end{lemma}

\begin{proof}
	By the definitions of the reference SGD iterate $g^{(T+1)}$ and the regularized solution
	\[
	g_{M,\lambda_t}
	=
	(L_M+\lambda_t I)^{-1}L_Mg_\rho,
	\]
	we obtain
	\begin{equation*}
		\begin{aligned}
			g^{(T+1)}-g_{M,\lambda_{T+1}}
			&=
			(1-\eta_T\lambda_T)g^{(T)}
			-\eta_T(g^{(T)}(x_T)-y_T)k_{M,x_T}
			-g_{M,\lambda_{T+1}} \\
			&=
			\bigl(I-\eta_T(L_M+\lambda_TI)\bigr)
			(g^{(T)}-g_{M,\lambda_T})
			+\eta_TL_Mg^{(T)}
			+g_{M,\lambda_T}   \\
			&\qquad
			-\eta_T(L_M+\lambda_TI)g_{M,\lambda_T}
			-\eta_T(g^{(T)}(x_T)-y_T)k_{M,x_T}
			-g_{M,\lambda_{T+1}} \\
			&=
			\bigl(I-\eta_T(L_M+\lambda_TI)\bigr)
			(g^{(T)}-g_{M,\lambda_T})
			-(g_{M,\lambda_{T+1}}-g_{M,\lambda_T}) \\
			&\qquad
			+\eta_TL_M(g^{(T)}-g_\rho)
			-\eta_T(g^{(T)}(x_T)-y_T)k_{M,x_T} \\
			&=
			\bigl(I-\eta_T(L_M+\lambda_TI)\bigr)
			(g^{(T)}-g_{M,\lambda_T})
			-(g_{M,\lambda_{T+1}}-g_{M,\lambda_T})
			+\eta_T\mathcal B_T,
		\end{aligned}
	\end{equation*}
	where
	\begin{equation}
		\label{eq:definition-Bt}
		\mathcal B_t
		=
		(y_t-g^{(t)}(x_t))k_{M,x_t}
		+
		L_M(g^{(t)}-g_\rho).
	\end{equation}
	
	Moreover,
	\[
	\mathbb E_{z_t}
	\!\left[
	\mathcal B_t
	\,\middle|\,
	z_1,\ldots,z_{t-1}
	\right]
	=
	0,
	\]
	that is, $\{\mathcal B_t\}_{t\ge1}$ forms a martingale difference sequence.
	
	Iterating the above recursion and using the initialization
	$g^{(1)}=0$, we obtain
	\begin{equation}
		\label{equ: middle term of reference SGD error}
		\begin{aligned}
			&\mathbb E_{Z^T}
			\!\left[
			\|g^{(T+1)}-g_{M,\lambda_{T+1}}\|_\rho^2
			\right]                                                   \\
			&\le
			3
			\left\|
			\prod_{t=1}^T
			\bigl(I-\eta_t(L_M+\lambda_tI)\bigr)
			g_{M,\lambda_1}
			\right\|_\rho^2                                            \\
			&\quad
			+
			3
			\left\|
			\sum_{t=1}^T
			\prod_{i=t+1}^T
			\bigl(I-\eta_i(L_M+\lambda_iI)\bigr)
			(g_{M,\lambda_{t+1}}-g_{M,\lambda_t})
			\right\|_\rho^2                                            \\
			&\quad
			+
			3
			\mathbb E_{Z^T}
			\left\|
			\sum_{t=1}^T
			\eta_t
			\prod_{i=t+1}^T
			\bigl(I-\eta_i(L_M+\lambda_iI)\bigr)
			\mathcal B_t
			\right\|_\rho^2,
		\end{aligned}
	\end{equation}
	where we have used the elementary inequality
	\[
	(a+b+c)^2
	\le
	3(a^2+b^2+c^2).
	\]
	
	Furthermore, by applying the martingale difference property, i.e., $\mathbb{E}_{z_t} [\mathcal{B}_t \mid z_1, \dots, z_{t-1}] = 0$ to the third term on the right-hand side of (\ref{equ: middle term of reference SGD error}), the cross terms vanish. Therefore,
	\begin{align*}
		& \mathbb{E}_{Z^{T}}\left[\left\|\sum_{t=1}^{T} \eta_{t} \prod_{i=t+1}^{T}\left(I-\eta_{i}\left(L_M+\lambda_i I\right)\right) \mathcal{B}_{t}\right\|_{\rho}^{2}\right] 
		= \sum_{t=1}^{T} \eta_t^2 \mathbb{E}_{Z^{t}}\left[ \left\| \prod_{i=t+1}^{T}\left(I-\eta_{i}\left(L_M+\lambda_i I\right)\right) \mathcal{B}_{t} \right\|_{\rho}^2\right]  \\
		&\qquad + \sum_{t=1}^{T}\sum_{s \neq t} \eta_t \eta_s \mathbb{E}_{Z^{T}} \left\langle \prod_{i=t+1}^{T}\left(I-\eta_{i}\left(L_M+\lambda_i I\right)\right) \mathcal{B}_{t}, \prod_{i=s+1}^{T}\left(I-\eta_{i}\left(L_M+\lambda_i I\right)\right) \mathcal{B}_{s} \right\rangle_{\rho} \\
		&= \sum_{t=1}^{T} \eta_t^2 \mathbb{E}_{Z^{t}}\left[ \left\| \prod_{i=t+1}^{T}\left(I-\eta_{i}\left(L_M+\lambda_i I\right)\right) \mathcal{B}_{t} \right\|_{\rho}^2\right] . 
	\end{align*}
	To bound each summand, recall that for every $f\in\mathcal{H}_M$,
	\[
	\|f\|_\rho^2
	=
	\left\|L_M^{1/2}f\right\|_{\mathcal{H}_M}^2.
	\]
	Since $\prod_{i=t+1}^{T}\left(I-\eta_{i}\left(L_M+\lambda_i I\right)\right) $ is a deterministic function of $L_M$, it commutes with
	$L_M$. Expanding the squared norm and taking the conditional expectation give	
	For $1 \le t \le T-1$, 
	we can derive 
	\begin{align*}
		&\qquad \mathbb{E}_{Z^{t}} \left[\left\|\prod_{i=t+1}^{T}\left(I-\eta_{i}\left(L_M+\lambda_i I\right)\right) \mathcal{B}_{t}\right\|_{\rho}^{2} \right]
		= \mathbb{E}_{Z^{t}} \left[\left\|L_M^{\frac{1}{2}} \prod_{i=t+1}^{T}\left(I-\eta_{i}\left(L_M+\lambda_i I\right)\right) \mathcal{B}_{t}\right\|_{\mathcal{H}_{M}}^{2}\right] \\
		&= \mathbb{E}_{Z^{t}} \left[\left\|L_M^{\frac{1}{2}} \prod_{i=t+1}^{T}\left(I-\eta_{i}\left(L_M+\lambda_i I\right)\right) \left( (y_t - g^{(t)}(x_t))k_{M, x_t} \right) \right\|_{\mathcal{H}_{M}}^{2} \right]\\
		&\quad + \mathbb{E}_{Z^{t-1}} \left[\left\|L_M^{\frac{1}{2}} \prod_{i=t+1}^{T}\left(I-\eta_{i}\left(L_M+\lambda_i I\right)\right) \left( L_M (g^{(t)} - g_{\rho}) \right) \right\|_{\mathcal{H}_{M}}^{2}\right] \\
		&\quad + 2\mathbb{E}_{Z^{t}} \left\langle L_M^{\frac{1}{2}} \prod_{i=t+1}^{T}\left(I-\eta_{i}\left(L_M+\lambda_i I\right)\right) \left( (y_t - g^{(t)}(x_t))k_{M, x_t} \right) , \right. \\
		&\qquad \qquad \quad \left. L_M^{\frac{1}{2}} \prod_{i=t+1}^{T}\left(I-\eta_{i}\left(L_M+\lambda_i I\right)\right)\left( L_M (g^{(t)} - g_{\rho}) \right) \right\rangle_{\mathcal{H}_{M}} \\
		&=\mathbb{E}_{Z^{t}} \left[\left\|L_M^{\frac{1}{2}} \prod_{i=t+1}^{T}\left(I-\eta_{i}\left(L_M+\lambda_i I\right)\right) \left( (y_t - g^{(t)}(x_t))k_{M, x_t} \right) \right\|_{\mathcal{H}_{M}}^{2}\right] \\
		&\quad - \mathbb{E}_{Z^{t-1}}\left[ \left\|L_M^{\frac{1}{2}} \prod_{i=t+1}^{T}\left(I-\eta_{i}\left(L_M+\lambda_i I\right)\right) \left( L_M (g^{(t)} - g_{\rho}) \right) \right\|_{\mathcal{H}_{M}}^{2} \right]\\
		&\leq \mathbb{E}_{Z^{t}} \left[\left\|L_M^{\frac{1}{2}} \prod_{i=t+1}^{T}\left(I-\eta_{i}\left(L_M+\lambda_i I\right)\right) \left( (y_t - g^{(t)}(x_t))k_{M, x_t} \right) \right\|_{\mathcal{H}_{M}}^{2}\right]. 
	\end{align*}
	Since 	 $|y_t|\leq1$, we have 
	\[
	\bigl(y_t-g^{(t)}(x_t)\bigr)^2
	\leq
	2\left(1+\|g^{(t)}\|_\infty^2\right).
	\]
	Moreover,
	 $\| f \|_{\mathcal{H}_M}^2 = \textnormal{Tr}(f \otimes_{\mathcal{H}_M} f)$, and
	\[
	\mathbb{E}_{x_t}
	\left[
	k_{M,x_t}\otimes_{\mathcal{H}_M}k_{M,x_t}
	\right]
	=
	L_M.
	\]
	Consequently,  
	\begin{align*}
		&\qquad \mathbb{E}_{Z^{t}} \left\|L_M^{\frac{1}{2}} \prod_{i=t+1}^{T}\left(I-\eta_{i}\left(L_M+\lambda_i I\right)\right) \left( (y_t - g^{(t)}(x_t))k_{M, x_t} \right) \right\|_{\mathcal{H}_{M}}^{2} \\
		&= \mathbb{E}_{Z^{t}} \textnormal{Tr} \left[ L_M  \prod_{i=t+1}^{T}\left(I-\eta_{i}\left(L_M+\lambda_i I\right)\right)^2 (y_t - g^{(t)}(x_t))^2 (k_{M, x_t} \otimes_{\mathcal{H}_M} k_{M, x_t}) \right] \\
		&\leq 2\left( 1 + \mathbb{E}_{Z^{t-1}} \left[\left\| g^{(t)} \right\|_{\infty}^2\right] \right) \textnormal{Tr} \left[ L_M^2 \prod_{i=t+1}^{T}\left(I-\eta_{i}\left(L_M+\lambda_i I\right)\right)^2 \right]. 
	\end{align*}
	Then, substituting the above estimates back into $(\ref{equ: middle term of reference SGD error})$ yields the desired result.
\end{proof}

To bound the reference SGD error $\mathbb{E}_{Z^{T}} \left[\left\|g^{(T+1)}-g_{M, \lambda_{T+1}}\right\|_{\rho}^{2}\right]$, we individually estimate the three terms on the right hand side of  (\ref{chap7:eq:error-decomp-of-convergence-error}), as shown in Propositions \ref{chap7:prop:first-term}, \ref{chap7:prop:second-term}, and \ref{chap7:prop:third-term}.
\begin{proposition}
	\label{chap7:prop:first-term}
	Let $\eta_t=\eta (t+1)^{-\theta}$ and $\lambda_t=1/\eta (t+1)^{-1+\theta}$  with $\frac12<\theta<1$ and $0<\eta<\frac{1-\theta}{4\kappa^2},$ and $g_{M, \lambda_1}=(\lambda_1 I+L_M)^{-1}L_M g_\rho$, we have 
	\begin{align}
		\label{chap7:eq:first-term}
		\left\|\prod_{t=1}^{T}\left(I-\eta_{t}(L_M + \lambda_t I)\right) g_{M, \lambda_1}\right\|_{\rho} \leq 
		\| g_{\rho} \|_{\rho} (T+1)^{-1}. 
	\end{align}
\end{proposition}
\begin{proof}
	By the definition of $g_{M,\lambda_1}$,
	\[
	g_{M,\lambda_1}
	=
	(L_M+\lambda_1I)^{-1}L_Mg_\rho,
	\]
	and the identity
	\[
	\eta_t\lambda_t=(t+1)^{-1},
	\]
	we have
	\begin{align*}		
		\left\|
		\prod_{t=1}^{T}
		\left(I-\eta_t(L_M+\lambda_tI)\right)
		g_{M,\lambda_1}
		\right\|_{\rho}
		&\le
		\prod_{t=1}^{T}
		\left\|
		I-\eta_t(L_M+\lambda_tI)
		\right\|_{\operatorname{op}}
		\,
		\|g_{M,\lambda_1}\|_{\rho}
		\\
		&\le
		\prod_{t=1}^{T}
		(1-\eta_t\lambda_t)
		\,
		\|g_\rho\|_{\rho}
		\\
		&=
		\prod_{t=1}^{T}
		\left(1-\frac{1}{t+1}\right)
		\|g_\rho\|_{\rho}
		=
		\frac{\|g_\rho\|_{\rho}}{T+1},
	\end{align*}
	where the second inequality follows from the facts that
	\[
	\|g_{M,\lambda_1}\|_{\rho}\le \|g_\rho\|_{\rho}
	\]
	and
	\[
	\left\|
	I-\eta_t(L_M+\lambda_tI)
	\right\|_{\operatorname{op}}
	\le
	1-\eta_t\lambda_t.
	\]
	This completes the proof.
\end{proof}

Before estimating the second term on the right-hand side of
\eqref{chap7:eq:error-decomp-of-convergence-error}, we establish the following
regularization-path estimate.
\begin{lemma}
	\label{chap7:lem:drift-lemma}
	Assume that Assumption \ref{assumption regularity condition} hold with $\frac12<r\le 1$. Then for any $\lambda, \mu \geq 0$, the following estimates hold:
	\begin{enumerate}
		\item[(1).] $\left\| g_{\infty, \lambda} - g_{\infty, \mu} \right\|_{\rho} \leq \frac{\left\| L_{\infty}^{-r} g_{\rho} \right\|_{\rho}}{r} \big| \lambda^{r} - \mu^{r} \big|$; 
		\item[(2).] $\left\| g_{\infty, \lambda} - g_{\infty, \mu} \right\|_{\mathcal{H}_{\infty}} \leq \frac{\left\| L_{\infty}^{-r} g_{\rho} \right\|_{\rho}}{r - \frac{1}{2}} \big| \lambda^{r-\frac{1}{2}} - \mu^{r-\frac{1}{2}} \big|$. 
	\end{enumerate}
\end{lemma}

\begin{proof}
		By symmetry, it suffices to consider $\lambda>\mu\geq0$. For $u>0$, define
	\[
	J_{u,\lambda,\mu}
	:=
	(\mu-\lambda)
	(L_\infty+\lambda I)^{-1}
	(L_\infty+\mu I)^{-1}
	L_\infty^{1+u}.
	\]
	
	We first establish an operator-norm estimate for
	$J_{u,\lambda,\mu}$. For $0<u\leq1$, the spectral theorem gives
	\begin{align}
		\|J_{u,\lambda,\mu}\|_{\operatorname{op}}
		=
		(\lambda-\mu)
		\sup_{s\geq0}
		\frac{s^{1+u}}{(s+\lambda)(s+\mu)}
\leq
		(\lambda-\mu)
		\sup_{s\geq0}
		\frac{s^u}{s+\lambda}
\leq
		(\lambda-\mu)\lambda^{u-1}.
		\label{equation: preliminary bound Ju}
	\end{align}
	Writing $a=\mu/\lambda\in[0,1)$, we have
	\[
	(\lambda-\mu)\lambda^{u-1}
	=
	(\lambda^u-\mu^u)
	\frac{1-a}{1-a^u}.
	\]
	Since
	\[
	u(1-a)\leq1-a^u,
	\qquad 0<u\leq1,\quad 0\leq a\leq1,
	\]
	it follows that
	\begin{equation}
		\label{J_u_lambda_mu}
		\|J_{u,\lambda,\mu}\|_{\operatorname{op}}
		\leq
		\frac{1}{u}
		\left|\lambda^u-\mu^u\right|.
	\end{equation}
	
	We now prove the two stated estimates.
	
	\begin{enumerate}
		\item
		From
		\[
		g_{\infty,\lambda}
		=
		(L_\infty+\lambda I)^{-1}L_\infty g_\rho
		\]
		and the resolvent identity
		\[
		(L_\infty+\lambda I)^{-1}
		-
		(L_\infty+\mu I)^{-1}
		=
		(\mu-\lambda)
		(L_\infty+\lambda I)^{-1}
		(L_\infty+\mu I)^{-1},
		\]
		we obtain
		\begin{align*}
			g_{\infty,\lambda}-g_{\infty,\mu}
			=
			(\mu-\lambda)
			(L_\infty+\lambda I)^{-1}
			(L_\infty+\mu I)^{-1}
			L_\infty g_\rho
=
			J_{r,\lambda,\mu}L_\infty^{-r}g_\rho.
		\end{align*}
		Hence, by \eqref{J_u_lambda_mu},
		\begin{align*}
			\left\|
			g_{\infty,\lambda}-g_{\infty,\mu}
			\right\|_\rho
			\leq
			\|J_{r,\lambda,\mu}\|_{\operatorname{op}}
			\|L_\infty^{-r}g_\rho\|_\rho
\leq
			\frac{\|L_\infty^{-r}g_\rho\|_\rho}{r}
			\left|\lambda^r-\mu^r\right|.
		\end{align*}
		
		\item
		Using the standard identification between
		$\mathcal H_\infty$ and the range of $L_\infty^{1/2}$, we have
		\[
		\|f\|_{\mathcal H_\infty}
		=
		\|L_\infty^{-1/2}f\|_\rho
		\]
		for every $f$ in the relevant range. Therefore,
		\begin{align*}
			\left\|
			g_{\infty,\lambda}-g_{\infty,\mu}
			\right\|_{\mathcal H_\infty}
			&=
			\left\|
			L_\infty^{-1/2}
			\left(
			g_{\infty,\lambda}-g_{\infty,\mu}
			\right)
			\right\|_\rho
			\\
			&=
			\left\|
			J_{r-\frac12,\lambda,\mu}
			L_\infty^{-r}g_\rho
			\right\|_\rho
			\\
			&\leq
			\|J_{r-\frac12,\lambda,\mu}\|_{\operatorname{op}}
			\|L_\infty^{-r}g_\rho\|_\rho.
		\end{align*}
		Since $0<r-\frac12\leq\frac12$, applying
		\eqref{J_u_lambda_mu} with $u=r-\frac12$ yields
		\begin{align*}
			\left\|
			g_{\infty,\lambda}-g_{\infty,\mu}
			\right\|_{\mathcal H_\infty}
			&\leq
			\frac{\|L_\infty^{-r}g_\rho\|_\rho}
			{r-\frac12}
			\left|
			\lambda^{r-\frac12}
			-
			\mu^{r-\frac12}
			\right|.
		\end{align*}
	\end{enumerate}
	The proof is complete.
\end{proof}
	The following proposition provides an upper bound for the second term in
\eqref{chap7:eq:error-decomp-of-convergence-error}, which measures the drift
induced by the time-varying regularization parameter.

\begin{proposition}\label{chap7:prop:second-term}
	Let $g_{M,\lambda_t}$ be defined by (\ref{equation: expression of g_M,lambda}) and
	\[
	\eta_t=\eta(t+1)^{-\theta},
	\qquad
	\lambda_t=\frac{1}{\eta}(t+1)^{-1+\theta},
	\]
	where $\frac12<\theta<1$ and $0<\eta<\frac{1-\theta}{4\kappa^2}$. If
	\[
	M\geq T^{2+2(r+1)(1-\theta)},
	\]
	then, with probability at least $1-\delta$,
	\begin{align}			
		\left\|
		\sum_{t=1}^{T}
		\prod_{i=t+1}^{T}
		\left(I-\eta_i(L_M+\lambda_i I)\right)
		\left(g_{M,\lambda_{t+1}}-g_{M,\lambda_t}\right)
		\right\|_\rho
		\leq
		V_1T^{-r(1-\theta)}
		\sqrt{2\log\frac{2}{\delta}},
		\label{chap7:eq:second-term}
	\end{align}
	where
	\[
	V_1
	=
	\kappa^2\eta 2^{3-\theta}\|g_\rho\|_\rho
	+
	\frac{
		2\eta^{-r}2^\theta(1-\theta)
		\|L_\infty^{-r}g_\rho\|_\rho
	}{
		1-r(1-\theta)
	}.
	\]
\end{proposition}

\begin{proof}
	We first decompose the difference between two consecutive regularized
	solutions as
	\begin{align*}
		g_{M,\lambda_{t+1}}-g_{M,\lambda_t}
		&=
		\left(g_{M,\lambda_{t+1}}-g_{\infty,\lambda_{t+1}}\right)
		+
		\left(g_{\infty,\lambda_{t+1}}-g_{\infty,\lambda_t}\right)+
		\left(g_{\infty,\lambda_t}-g_{M,\lambda_t}\right).
	\end{align*}
	Consequently,
	\begin{align}
		&
		\left\|
		\sum_{t=1}^{T}
		\prod_{i=t+1}^{T}
		\left(I-\eta_i(L_M+\lambda_i I)\right)
		\left(g_{M,\lambda_{t+1}}-g_{M,\lambda_t}\right)
		\right\|_\rho
		\nonumber\\
		&\leq
		\sum_{t=1}^{T}
		\left\|
		g_{M,\lambda_{t+1}}-g_{\infty,\lambda_{t+1}}
		\right\|_\rho
		+
		\sum_{t=1}^{T}
		\left\|
		g_{\infty,\lambda_t}-g_{M,\lambda_t}
		\right\|_\rho
		\nonumber\\
		&\quad+
		\sum_{t=1}^{T}
		\left\|
		\prod_{i=t+1}^{T}
		\left(I-\eta_i(L_M+\lambda_i I)\right)
		\right\|_{\operatorname{op}}
		\left\|
		g_{\infty,\lambda_{t+1}}-g_{\infty,\lambda_t}
		\right\|_\rho
		\nonumber\\
		&\leq
		2\sum_{t=1}^{T+1}
		\left\|
		g_{M,\lambda_t}-g_{\infty,\lambda_t}
		\right\|_\rho+
		\sum_{t=1}^{T}
		\left\|
		\prod_{i=t+1}^{T}
		\left(I-\eta_i(L_M+\lambda_i I)\right)
		\right\|_{\operatorname{op}}
		\left\|
		g_{\infty,\lambda_{t+1}}-g_{\infty,\lambda_t}
		\right\|_\rho.
		\label{chap7:eq:drift-error-decomp}
	\end{align}
	We estimate the two terms on the right-hand side separately.
	
	\paragraph{Finite-width approximation error.}
	By the definitions of $g_{M,\lambda_t}$ and $g_{\infty,\lambda_t}$,
	\begin{align*}
		g_{M,\lambda_t}-g_{\infty,\lambda_t}
		&=
		\left[
		(L_M+\lambda_t I)^{-1}L_M
		-
		(L_\infty+\lambda_t I)^{-1}L_\infty
		\right]g_\rho
		\\
		&=
		\lambda_t
		\left[
		(L_\infty+\lambda_t I)^{-1}
		-
		(L_M+\lambda_t I)^{-1}
		\right]g_\rho
		\\
		&=
		\lambda_t
		(L_\infty+\lambda_t I)^{-1}
		(L_M-L_\infty)
		(L_M+\lambda_t I)^{-1}g_\rho,
	\end{align*}
	where the last equality follows from the resolvent identity.
	
	Since
	$
	\left\|
	\lambda_t(L_\infty+\lambda_t I)^{-1}
	\right\|_{\operatorname{op}}
	\leq1
	$
	and
	$
	\left\|
	(L_M+\lambda_t I)^{-1}
	\right\|_{\operatorname{op}}
	\leq\lambda_t^{-1},
	$
	we obtain
	\begin{align}
		\sum_{t=1}^{T+1}
		\left\|
		g_{M,\lambda_t}-g_{\infty,\lambda_t}
		\right\|_\rho
		&\leq
		\|L_M-L_\infty\|_{\operatorname{op}}
		\|g_\rho\|_\rho
		\sum_{t=1}^{T+1}\lambda_t^{-1}
		\nonumber\\
		&=
		\eta
		\|L_M-L_\infty\|_{\operatorname{op}}
		\|g_\rho\|_\rho
		\sum_{t=1}^{T+1}(t+1)^{1-\theta}
		\nonumber\\
		&\leq
		\eta(T+2)^{2-\theta}
		\|L_M-L_\infty\|_{\operatorname{op}}
		\|g_\rho\|_\rho
		\nonumber\\
		&\leq
		\eta 2^{2-\theta}T^{2-\theta}
		\|L_M-L_\infty\|_{\operatorname{op}}
		\|g_\rho\|_\rho.
		\label{difference of g_Mlambda and g_inftylambda}
	\end{align}
	
	By part~(a) of Lemma~\ref{lemmaA}, with probability at least $1-\delta$,
	\[
	\|L_M-L_\infty\|_{\operatorname{op}}
	\leq
	\kappa^2
	\sqrt{\frac{2\log(2/\delta)}{M}}.
	\]
	Under the width condition
	\[
	M\geq T^{2+2(r+1)(1-\theta)},
	\]
	this gives
	\begin{align*}
		\|L_M-L_\infty\|_{\operatorname{op}}
		&\leq
		\kappa^2
		\sqrt{2\log\frac{2}{\delta}}\,
		T^{-1-(r+1)(1-\theta)}.
	\end{align*}
	Substituting this bound into
	\eqref{difference of g_Mlambda and g_inftylambda}, and using
	$
	2-\theta-1-(r+1)(1-\theta)
	=
	-r(1-\theta),
	$
	we obtain, with probability at least $1-\delta$,
	\begin{align}
		\sum_{t=1}^{T+1}
		\left\|
		g_{M,\lambda_t}-g_{\infty,\lambda_t}
		\right\|_\rho
		&\leq
		\kappa^2\eta2^{2-\theta}\|g_\rho\|_\rho
		\sqrt{2\log\frac{2}{\delta}}\,
		T^{-r(1-\theta)}.
		\label{chap7:eq:drift-error-decomp-first-term}
	\end{align}
	
	\paragraph{Regularization-path drift.}
	By part~(1) of Lemma~\ref{chap7:lem:drift-lemma},
	\begin{align*}
		\left\|
		g_{\infty,\lambda_{t+1}}-g_{\infty,\lambda_t}
		\right\|_\rho
		&\leq
		\frac{\|L_\infty^{-r}g_\rho\|_\rho}{r}
		\left|
		\lambda_{t+1}^r-\lambda_t^r
		\right|
		\\
		&=
		\frac{\eta^{-r}\|L_\infty^{-r}g_\rho\|_\rho}{r}
		\left|
		(t+2)^{-r(1-\theta)}
		-
		(t+1)^{-r(1-\theta)}
		\right|.
	\end{align*}
	By the mean value theorem,
	\[
	\left|
	(t+2)^{-r(1-\theta)}
	-
	(t+1)^{-r(1-\theta)}
	\right|
	\leq
	r(1-\theta)(t+1)^{-r(1-\theta)-1}.
	\]
	Therefore,
	\begin{equation}
		\label{equation: regularization increment}
		\left\|
		g_{\infty,\lambda_{t+1}}-g_{\infty,\lambda_t}
		\right\|_\rho
		\leq
		\eta^{-r}(1-\theta)
		\|L_\infty^{-r}g_\rho\|_\rho
		(t+1)^{-r(1-\theta)-1}.
	\end{equation}
	
	Since $\eta_i\lambda_i=(i+1)^{-1}$ and
	\[
	\left\|
	I-\eta_i(L_M+\lambda_i I)
	\right\|_{\operatorname{op}}
	\leq
	1-\eta_i\lambda_i,
	\]
	for $1\le t\le T-1,$ we have
	\begin{align}
		\left\|
		\prod_{i=t+1}^{T}
		\left(I-\eta_i(L_M+\lambda_i I)\right)
		\right\|_{\operatorname{op}}
		\leq
		\prod_{i=t+1}^{T}
		\left(1-\frac{1}{i+1}\right)
		=
		\prod_{i=t+1}^{T}\frac{i}{i+1}
		=
		\frac{t+1}{T+1}.
		\label{equation: drift operator product}
	\end{align}
	Combining \eqref{equation: regularization increment} and
	\eqref{equation: drift operator product} gives
	\begin{align*}
		&
		\sum_{t=1}^{T-1}
		\left\|
		\prod_{i=t+1}^{T}
		\left(I-\eta_i(L_M+\lambda_i I)\right)
		\right\|_{\operatorname{op}}
		\left\|
		g_{\infty,\lambda_{t+1}}-g_{\infty,\lambda_t}
		\right\|_\rho
		\\
		&\leq
		\frac{
			\eta^{-r}(1-\theta)\|L_\infty^{-r}g_\rho\|_\rho
		}{T+1}
		\sum_{t=1}^{T}
		(t+1)^{-r(1-\theta)}.
	\end{align*}
	Since $0<r(1-\theta)<1$,
	\begin{align*}
		\sum_{t=1}^{T}
		(t+1)^{-r(1-\theta)}
		\leq
		\int_{1}^{T+1}x^{-r(1-\theta)}\,dx
\leq
		\frac{(T+1)^{1-r(1-\theta)}}{1-r(1-\theta)}.
	\end{align*}
	For $t=T,$ since $\prod_{i=t+1}^{T}
	\left(I-\eta_i(L_M+\lambda_i I)\right)
	=I$, and by (\ref{equation: regularization increment}), we have  
	$$	\left\|
	g_{\infty,\lambda_{T+1}}-g_{\infty,\lambda_T}
	\right\|_\rho
	\leq
	\eta^{-r}(1-\theta)
	\|L_\infty^{-r}g_\rho\|_\rho
	(T+1)^{-r(1-\theta)-1}.
	$$
	It follows that
	\begin{align}
		&
		\sum_{t=1}^{T}
		\left\|
		\prod_{i=t+1}^{T}
		\left(I-\eta_i(L_M+\lambda_i I)\right)
		\right\|_{\operatorname{op}}
		\left\|
		g_{\infty,\lambda_{t+1}}-g_{\infty,\lambda_t}
		\right\|_\rho
		\nonumber\\
		&	=	\sum_{t=1}^{T-1}
		\left\|
		\prod_{i=t+1}^{T}
		\left(I-\eta_i(L_M+\lambda_i I)\right)
		\right\|_{\operatorname{op}}
		\left\|
		g_{\infty,\lambda_{t+1}}-g_{\infty,\lambda_t}
		\right\|_\rho+\left\|
		g_{\infty,\lambda_{T+1}}-g_{\infty,\lambda_T}
		\right\|_\rho
		\nonumber\\
		&\leq
		\frac{
			\eta^{-r}(1-\theta)\|L_\infty^{-r}g_\rho\|_\rho
		}{
			1-r(1-\theta)
		}
		(T+1)^{-r(1-\theta)}+\eta^{-r}(1-\theta)
		\|L_\infty^{-r}g_\rho\|_\rho
		(T+1)^{-r(1-\theta)-1}
		\nonumber\\
		&\leq
		\frac{
			2\eta^{-r}(1-\theta)\|L_\infty^{-r}g_\rho\|_\rho
		}{
			1-r(1-\theta)
		}
		T^{-r(1-\theta)}.
		\label{equation:chap7:eq:drift-error-decomp-second-term}
	\end{align}
	
	Finally, substituting
	\eqref{chap7:eq:drift-error-decomp-first-term} and
	\eqref{equation:chap7:eq:drift-error-decomp-second-term}
	into \eqref{chap7:eq:drift-error-decomp} yields the desired estimate.
\end{proof}

Before estimating the third term in (\ref{chap7:eq:error-decomp-of-convergence-error}), we need the following lemma.

\begin{proposition}\label{prop: uniform bound for E(g(t))}
	Let $\{g^{(t)}\}_{t\geq 1}$ be generated by Algorithm~\ref{R-SGD}. 
	Suppose that
	\[
	\eta_k=\eta(k+1)^{-\theta},
	\qquad
	\lambda_k=\frac{1}{\eta}(k+1)^{-(1-\theta)},
	\]
	where $\frac{1}{2}<\theta<1$ and
	\begin{equation}\label{eq: conditon2 on eta}
	0<\eta\leq
	\sqrt{\frac{2\theta-1}{60\kappa^4\theta}}.
	\end{equation}
	Then, for every $t\geq 1$,
	\begin{equation}
		\label{eq:uniform-risk-bound}
		\mathbb{E}_{Z^{t-1}}
		\left[
		\mathcal{E}\bigl(g^{(t)}\bigr)
		\right]
		\leq
		40\|g_{\rho}\|_{\rho}^{2}
		+
		2\mathcal{E}(g_{\rho}).
	\end{equation}
\end{proposition}

\begin{proof}
	We prove \eqref{eq:uniform-risk-bound} by induction on $t$.
	
	For the base case, i.e., $t=1$, Algorithm~\ref{R-SGD} gives $g^{(1)}=0$. Therefore,
	\[
	\mathcal{E}(g^{(1)})
	= 	\mathcal{E}(g^{(1)})-\mathcal{E}(g_\rho) +\mathcal{E}(g_\rho)= 	\|g^{(1)}-g_\rho\|_\rho^2+\mathcal{E}(g_\rho)=
	\|0-g_\rho\|_\rho^2+\mathcal{E}(g_\rho)
	\leq
	40\|g_\rho\|_\rho^2+2\mathcal{E}(g_\rho).
	\]
	
	Assume that \eqref{eq:uniform-risk-bound} holds for every
	$1\leq k\leq t$. We show that it also holds for $t+1$.
	
	By the definition of the excess risk,
	\begin{equation}
		\label{eq:risk-decomposition-induction}
		\begin{aligned}
			\mathcal{E}\bigl(g^{(t+1)}\bigr)
			&=
			\mathcal{E}\bigl(g^{(t+1)}\bigr)
			-\mathcal{E}(g_{\rho})
			+\mathcal{E}(g_{\rho}) \\
			&=
			\left\|g^{(t+1)}-g_{\rho}\right\|_{\rho}^{2}
			+\mathcal{E}(g_{\rho}) \\
			&\leq
			2\left\|g^{(t+1)}-g_{M,\lambda_{t+1}}\right\|_{\rho}^{2}
			+
			2\left\|g_{M,\lambda_{t+1}}-g_{\rho}\right\|_{\rho}^{2}
			+\mathcal{E}(g_{\rho}).
		\end{aligned}
	\end{equation}
	
	Since
	\[
	g_{M,\lambda_{t+1}}
	=
	(L_M+\lambda_{t+1}I)^{-1}L_Mg_{\rho},
	\]
	we have
	\begin{equation}
		\label{eq:regularization-bias-bound}
		\begin{aligned}
			\left\|g_{M,\lambda_{t+1}}-g_{\rho}\right\|_{\rho}^{2}
			&=
			\left\|
			(L_M+\lambda_{t+1}I)^{-1}L_Mg_{\rho}
			-g_{\rho}
			\right\|_{\rho}^{2} \\
			&= \left\|
			\lambda_{t+1}(L_M+\lambda_{t+1}I)^{-1}g_{\rho}
			\right\|_{\rho}^{2} \\
			&\leq \left\|
			\lambda(L_M+\lambda I)^{-1}
			\right\|_{\mathrm{op}}^2\cdot \|g_{\rho}\|_{\rho}^{2}
			\leq
			\|g_{\rho}\|_{\rho}^{2}.
		\end{aligned}
	\end{equation}
	Therefore, it suffices to bound
	\[
	\mathbb{E}_{Z^t}
	\left[
	\left\|g^{(t+1)}-g_{M,\lambda_{t+1}}\right\|_{\rho}^{2}
	\right].
	\]
	
	By \eqref{equ: middle term of reference SGD error}, established in the
	proof of Lemma~\ref{chap7:lem:error-decomp-of-convergence-error}, together with the martingale-difference
	property of $\mathcal{B}_i$, we obtain
	\begin{equation}
		\label{eq:three-term-decomposition}
		\begin{aligned}
			\mathbb{E}_{Z^t}
			\left[
			\left\|g^{(t+1)}-g_{M,\lambda_{t+1}}\right\|_{\rho}^{2}
			\right] 
			&\quad\leq
			3\left\|
			\prod_{i=1}^{t}
			\left(I-\eta_i(L_M+\lambda_i I)\right)
			g_{M,\lambda_1}
			\right\|_{\rho}^{2} \\
			&\qquad+
			3\left\|
			\sum_{i=1}^{t}
			\prod_{k=i+1}^{t}
			\left(I-\eta_k(L_M+\lambda_k I)\right)
			\bigl(g_{M,\lambda_{i+1}}-g_{M,\lambda_i}\bigr)
			\right\|_{\rho}^{2} \\
			&\qquad+
			3\sum_{i=1}^{t}
			\eta_i^2
			\mathbb{E}_{Z^i}
			\left[
			\left\|
			\prod_{k=i+1}^{t}
			\left(I-\eta_k(L_M+\lambda_k I)\right)
			\mathcal{B}_i
			\right\|_{\rho}^{2}
			\right].
		\end{aligned}
	\end{equation}
	
	We first bound the first term on the right-hand side of
	\eqref{eq:three-term-decomposition}. Using
	\[
	g_{M,\lambda}
	=
	(L_M+\lambda I)^{-1}L_Mg_{\rho}
	\]
	and the contraction property of the operator products, we obtain
	\begin{equation}
		\label{eq:initial-term-bound}
		\begin{aligned}
			\left\|
			\prod_{i=1}^{t}
			\left(I-\eta_i(L_M+\lambda_i I)\right)
			g_{M,\lambda_1}
			\right\|_{\rho}^{2}
			\leq
			\left\|g_{M,\lambda_1}\right\|_{\rho}^{2} 
			=
			\left\|
			(L_M+\lambda_1I)^{-1}L_Mg_{\rho}
			\right\|_{\rho}^{2} 
			\leq
			\|g_{\rho}\|_{\rho}^{2}.
		\end{aligned}
	\end{equation}
	
	We next bound the second term on the right-hand side of
	\eqref{eq:three-term-decomposition}. By the resolvent identity,
	\begin{equation}
		\label{eq:regularized-solution-difference}
		\begin{aligned}
			\left\|g_{M,\lambda_k}-g_{M,\lambda_{k+1}}\right\|_{\rho}
			&=
			\left\|
			\left[
			(L_M+\lambda_kI)^{-1}
			-
			(L_M+\lambda_{k+1}I)^{-1}
			\right]
			L_Mg_{\rho}
			\right\|_{\rho} \\
			&=
			\left\|
			(L_M+\lambda_kI)^{-1}
			(\lambda_{k+1}-\lambda_k)
			(L_M+\lambda_{k+1}I)^{-1}
			L_Mg_{\rho}
			\right\|_{\rho} \\
			&\leq
			\lambda_k^{-1}
			(\lambda_k-\lambda_{k+1})
			\|g_{\rho}\|_{\rho}.
		\end{aligned}
	\end{equation}
	Consequently,
	\begin{equation}
		\label{eq:path-difference-preliminary}
		\begin{aligned}
			&\left\|
			\sum_{k=1}^{t}
			\prod_{j=k+1}^{t}
			\left(I-\eta_j(L_M+\lambda_jI)\right)
			\bigl(g_{M,\lambda_{k+1}}-g_{M,\lambda_k}\bigr)
			\right\|_{\rho} \\
			&\quad\leq
			\sum_{k=1}^{t-1}
			\left\|
			\prod_{j=k+1}^{t}
			\left(I-\eta_j(L_M+\lambda_jI)\right)
			\right\|_{\mathrm{op}}
			\left\|g_{M,\lambda_{k+1}}-g_{M,\lambda_k}\right\|_{\rho} +
			\left\|g_{M,\lambda_{t+1}}-g_{M,\lambda_t}\right\|_{\rho} \\
			&\quad\leq
			\left[
			\sum_{k=1}^{t-1}
			\prod_{j=k+1}^{t}(1-\eta_j\lambda_j)
			\lambda_k^{-1}(\lambda_k-\lambda_{k+1})
			+
			\lambda_t^{-1}(\lambda_t-\lambda_{t+1})
			\right]
			\|g_{\rho}\|_{\rho}.
		\end{aligned}
	\end{equation}
	
	Under the prescribed choices of $\eta_j$ and $\lambda_j$, we have
	\[
	\eta_j\lambda_j=\frac{1}{j+1}.
	\]
	It follows that
	\[
	\prod_{j=k+1}^{t}(1-\eta_j\lambda_j)
	=
	\prod_{j=k+1}^{t}\frac{j}{j+1}
	=
	\frac{k+1}{t+1}.
	\]
	Therefore,
	\begin{equation}
		\label{eq:path-sum-expression}
		\begin{aligned}
			\sum_{k=1}^{t-1}
			\prod_{j=k+1}^{t}(1-\eta_j\lambda_j)
			\lambda_k^{-1}(\lambda_k-\lambda_{k+1}) 
			&=
			\sum_{k=1}^{t-1}
			\frac{k+1}{t+1}
			\lambda_k^{-1}(\lambda_k-\lambda_{k+1}) \\
			&=
			\frac{1}{t+1}
			\sum_{k=1}^{t-1}
			(k+1)
			\left[
			1-
			\left(\frac{k+1}{k+2}\right)^{1-\theta}
			\right].
		\end{aligned}
	\end{equation}
	
	To bound the last expression, we apply the mean value theorem to
	$h(x)=x^{1-\theta}$ over the interval
	$\left[(k+1)/(k+2),1\right]$. Since
	\[
	h'(x)=(1-\theta)x^{-\theta},
	\]
	there exists
	\[
	\xi_k\in\left(\frac{k+1}{k+2},1\right)
	\]
	such that
	\begin{equation}
		\label{eq:mvt-path-bound}
		\begin{aligned}
			1-\left(\frac{k+1}{k+2}\right)^{1-\theta}
			=
			(1-\theta)\xi_k^{-\theta}
			\left(1-\frac{k+1}{k+2}\right) 
			&\leq
			(1-\theta)
			\left(\frac{k+1}{k+2}\right)^{-\theta}
			\frac{1}{k+2} 
			\leq
			\frac{1-\theta}{k+1}.
		\end{aligned}
	\end{equation}
	Substituting \eqref{eq:mvt-path-bound} into
	\eqref{eq:path-sum-expression} gives
	\begin{equation}
		\label{eq:path-sum-bound}
		\begin{aligned}
			\sum_{k=1}^{t-1}
			\prod_{j=k+1}^{t}(1-\eta_j\lambda_j)
			\lambda_k^{-1}(\lambda_k-\lambda_{k+1}) 
			\leq
			\frac{1}{t+1}
			\sum_{k=1}^{t-1}(1-\theta)
			\leq
			1-\theta
			<1.
		\end{aligned}
	\end{equation}
	Moreover,
	\begin{equation}
		\label{eq:last-path-increment}
		\lambda_t^{-1}(\lambda_t-\lambda_{t+1})
		=
		1-\left(\frac{t+1}{t+2}\right)^{1-\theta}
		\leq 1.
	\end{equation}
	Combining
	\eqref{eq:path-difference-preliminary},
	\eqref{eq:path-sum-bound}, and
	\eqref{eq:last-path-increment}, we obtain
	\begin{equation}
		\label{eq:path-difference-final}
		\left\|
		\sum_{k=1}^{t}
		\prod_{j=k+1}^{t}
		\left(I-\eta_j(L_M+\lambda_jI)\right)
		\bigl(g_{M,\lambda_{k+1}}-g_{M,\lambda_k}\bigr)
		\right\|_{\rho}
		\leq
		2\|g_{\rho}\|_{\rho}.
	\end{equation}
	
	Finally, we bound the stochastic term on the right-hand side of
	\eqref{eq:three-term-decomposition}. Recall that
	\[
	\mathcal{B}_k
	=
	\bigl(y_k-g^{(k)}(x_k)\bigr)k_{M,x_k}
	+
	L_M\bigl(g^{(k)}-g_{\rho}\bigr)
	\].
	%
	Let
	\[
	\xi_k=
	\bigl(y_k-g^{(k)}(x_k)\bigr)k_{M,x_k}.
	\]
	By the definition of $\mathcal{B}_k$,
	\[
	\mathcal{B}_k
	=
	\xi_k-\mathbb{E}_{z_k}[\xi_k\mid z_1,\ldots,z_{k-1}],
	\]
	and hence
	\[
	\mathbb{E}_{z_k}[\mathcal{B}_k\mid z_1,\ldots,z_{k-1}]=0.
	\]
	The conditional variance inequality therefore gives
	\[
	\begin{aligned}
		\mathbb{E}_{z_k}
		\left[
		\|L_M^{1/2}\mathcal{B}_k\|_{\mathcal{H}_M}^2
		\mid z_1,\ldots,z_{k-1}
		\right]
		&\leq
		\mathbb{E}_{z_k}
		\left[
		\|L_M^{1/2}\xi_k\|_{\mathcal{H}_M}^2
		\mid z_1,\ldots,z_{k-1}
		\right] \\
		&\leq
		\kappa^4
		\mathbb{E}_{z_k}
		\left[
		\bigl(y_k-g^{(k)}(x_k)\bigr)^2
		\mid z_1,\ldots,z_{k-1}\right]\\
		&= \kappa^4
		\mathcal{E}\bigl(g^{(k)}\bigr)
	\end{aligned}
	\]
	we obtain
	\begin{equation}
		\label{eq:stochastic-term-bound}
		\begin{aligned}
			&\sum_{k=1}^{t}
			\eta_k^{2}
			\mathbb{E}_{Z^k}
			\left[
			\left\|
			\prod_{j=k+1}^{t}
			\left(I-\eta_j(L_M+\lambda_jI)\right)
			\mathcal{B}_k
			\right\|_{\rho}^{2}
			\right] \\
			&\quad\leq
			\sum_{k=1}^{t}
			\eta_k^{2}
			\mathbb{E}_{Z^k}
			\left[
			\|\mathcal{B}_k\|_{\rho}^{2}
			\right] \\
			&\quad=
			\sum_{k=1}^{t}
			\eta_k^{2}
			\mathbb{E}_{Z^k}
			\left[
			\left\|L_M^{1/2}\mathcal{B}_k\right\|_{\mathcal{H}_M}^{2}
			\right] \\
			&\quad\leq
			\kappa^4
			\sum_{k=1}^{t}
			\eta_k^{2}
			\mathbb{E}_{Z^{k-1}}
			\left[
			\mathcal{E}\bigl(g^{(k)}\bigr)
			\right].
		\end{aligned}
	\end{equation}
	
	By the induction hypothesis,
	\begin{equation}
		\label{eq:stochastic-term-induction}
		\begin{aligned}
			\kappa^4
			\sum_{k=1}^{t}
			\eta_k^{2}
			\mathbb{E}_{Z^{k-1}}
			\left[
			\mathcal{E}\bigl(g^{(k)}\bigr)
			\right] 
			\leq
			\kappa^4
			\left(
			40\|g_{\rho}\|_{\rho}^{2}
			+
			2\mathcal{E}(g_{\rho})
			\right)
			\sum_{k=1}^{t}\eta_k^{2}.
		\end{aligned}
	\end{equation}
	
	Since $\eta_k=\eta(k+1)^{-\theta}$ with $\theta>\frac12$, the integral
	test yields
	\begin{equation}
		\label{eq:step-size-square-sum}
		\begin{aligned}
			\sum_{k=1}^{t}\eta_k^{2}
			=
			\eta^{2}
			\sum_{k=1}^{t}(k+1)^{-2\theta} 
			\leq
			\eta^{2}
			\left(
			1+\int_{1}^{\infty}x^{-2\theta}\,dx
			\right) 
			=
			\eta^{2}\frac{2\theta}{2\theta-1}.
		\end{aligned}
	\end{equation}
	Under the condition
	\[
	\eta\leq
	\sqrt{\frac{2\theta-1}{60\kappa^4\theta}},
	\]
	we have
	\begin{equation}
		\label{eq:step-size-absorption}
		3\kappa^4
		\sum_{k=1}^{t}\eta_k^{2}
		\leq
		3\kappa^4\eta^{2}
		\frac{2\theta}{2\theta-1}
		\leq
		\frac{1}{10}.
	\end{equation}
	
	Combining
	\eqref{eq:three-term-decomposition},
	\eqref{eq:initial-term-bound},
	\eqref{eq:path-difference-final},
	\eqref{eq:stochastic-term-induction}, and
	\eqref{eq:step-size-absorption}, we obtain
	\begin{equation}
		\label{eq:induction-estimator-bound}
		\begin{aligned}
			\mathbb{E}_{Z^t}
			\left[
			\left\|g^{(t+1)}-g_{M,\lambda_{t+1}}\right\|_{\rho}^{2}
			\right] 
			\leq
			15\|g_{\rho}\|_{\rho}^{2}
			+
			\frac{1}{10}
			\left(
			40\|g_{\rho}\|_{\rho}^{2}
			+
			2\mathcal{E}(g_{\rho})
			\right).
		\end{aligned}
	\end{equation}
	
	Finally, substituting
	\eqref{eq:regularization-bias-bound} and
	\eqref{eq:induction-estimator-bound}
	into \eqref{eq:risk-decomposition-induction} and simplifying the
	constants yield
	\[
	\mathbb{E}_{Z^t}
	\left[
	\mathcal{E}\bigl(g^{(t+1)}\bigr)
	\right]
	\leq
	40\|g_{\rho}\|_{\rho}^{2}
	+
	2\mathcal{E}(g_{\rho}).
	\]
	This completes the induction and proves
	\eqref{eq:uniform-risk-bound}.
\end{proof}

\begin{proposition}
	\label{chap7:lem:uniform-boundedness-of-g_t}
	Suppose that Assumptions
	\ref{assumption activation function conditions},
	\ref{assumption uniform bounded}, and
	\ref{assumption regularity condition} hold. Let
	\[
	\eta_t=\eta(t+1)^{-\theta},
	\qquad
	\lambda_t=\frac{1}{\eta}(t+1)^{-(1-\theta)},
	\]
	where $\frac12<\theta<1$ and
	\[
	0<\eta
	\leq
	\sqrt{\frac{2\theta-1}{60\kappa^4\theta}}.
	\]
	If
	\[
	M\geq T^{2(2-\theta)},
	\]
	then, with probability at least $1-\delta$, for every
	$t\in\{1,\ldots,T+1\}$,
	\begin{align*}
		\mathbb{E}_{Z^{t-1}}
		\left[
		\left\|g^{(t)}\right\|_\infty^2
		\right]
		&\leq
		\Bigg[
		\left(
		8+
		\frac{256\kappa^4\eta^2}
		{2^{2\theta}\theta^2}
		\right)
		\kappa^{4r}
		\left\|L_\infty^{-r}g_\rho\right\|_\rho^2+
		\frac{
			192\kappa^2\eta^{1-2r}
			\left\|L_\infty^{-r}g_\rho\right\|_\rho^2
		}{
			(2r-1)^2
			\left(2-(2r-1)(1-\theta)\right)^2
		}
		\\
		&\quad+
		\frac{2}{5\kappa^2}
		\left(
		20\|g_\rho\|_\rho^2
		+
		\mathcal{E}(g_\rho)
		\right)
		\Bigg]
		\log\frac{2}{\delta}.
	\end{align*}
\end{proposition}

\begin{proof}
	We begin with the decomposition
	\begin{align}
		\mathbb{E}_{Z^{t-1}}
		\left[
		\left\|g^{(t)}\right\|_\infty^2
		\right]
		&\leq
		2\mathbb{E}_{Z^{t-1}}
		\left[
		\left\|g^{(t)}-g_{M,\lambda_t}\right\|_\infty^2
		\right]
		+
		2\left\|g_{M,\lambda_t}\right\|_\infty^2
		\nonumber\\
		&\leq
		2\mathbb{E}_{Z^{t-1}}
		\left[
		\left\|g^{(t)}-g_{M,\lambda_t}\right\|_\infty^2
		\right]
		+
		2\|g_\rho\|_\infty^2.
		\label{chap7:eq:g_t-decomp}
	\end{align}
	
	We first estimate the first term on the right-hand side of
	\eqref{chap7:eq:g_t-decomp}. By the error decomposition derived in the proof
	of Lemma~\ref{chap7:lem:error-decomp-of-convergence-error}, the reproducing
	property
	\[
	\|f\|_\infty
	\leq
	\kappa\|f\|_{\mathcal{H}_M},
	\qquad f\in\mathcal{H}_M,
	\]
	and the martingale-difference property of $\mathcal{B}_i$, we obtain
	\begin{align}		
		\mathbb{E}_{Z^{t-1}}
		\left[
		\left\|g^{(t)}-g_{M,\lambda_t}\right\|_\infty^2
		\right]
		&\leq
		3
		\left\|
		\prod_{i=1}^{t-1}
		\left(I-\eta_i(L_M+\lambda_iI)\right)
		g_{M,\lambda_1}
		\right\|_\infty^2
		\nonumber\\
		&\quad+
		3
		\left\|
		\sum_{i=1}^{t-1}
		\prod_{j=i+1}^{t-1}
		\left(I-\eta_j(L_M+\lambda_jI)\right)
		\left(g_{M,\lambda_{i+1}}-g_{M,\lambda_i}\right)
		\right\|_\infty^2
		\nonumber\\
		&\quad+
		3\mathbb{E}_{Z^{t-1}}
		\left[
		\left\|
		\sum_{i=1}^{t-1}
		\eta_i
		\prod_{j=i+1}^{t-1}
		\left(I-\eta_j(L_M+\lambda_jI)\right)
		\mathcal{B}_i
		\right\|_\infty^2
		\right]
		\nonumber\\
		&\leq
		3\|g_\rho\|_\infty^2
		+
		3
		\left\|
		\sum_{i=1}^{t-1}
		\prod_{j=i+1}^{t-1}
		\left(I-\eta_j(L_M+\lambda_jI)\right)
		\left(g_{M,\lambda_{i+1}}-g_{M,\lambda_i}\right)
		\right\|_\infty^2
		\nonumber\\
		&\quad+
		3\kappa^2
		\sum_{i=1}^{t-1}
		\eta_i^2
		\mathbb{E}_{Z^i}
		\left[
		\|\mathcal{B}_i\|_{\mathcal{H}_M}^2
		\right].
		\label{chap7:eq:temp-first-term-of-g_t-decomp}
	\end{align}
	
	We next estimate the drift term in
	\eqref{chap7:eq:temp-first-term-of-g_t-decomp}. Proceeding as in the proof of
	Proposition~\ref{chap7:prop:second-term}, we have
	\begin{align}
		&
		\left\|
		\sum_{i=1}^{t-1}
		\prod_{j=i+1}^{t-1}
		\left(I-\eta_j(L_M+\lambda_jI)\right)
		\left(g_{M,\lambda_{i+1}}-g_{M,\lambda_i}\right)
		\right\|_\infty
		\nonumber\\
		&\leq
		2\sum_{i=1}^{t}
		\left\|
		g_{\infty,\lambda_i}-g_{M,\lambda_i}
		\right\|_\infty+
		\sum_{i=1}^{t-1}
		\left\|
		\prod_{j=i+1}^{t-1}
		\left(I-\eta_j(L_M+\lambda_jI)\right)
		\right\|_{\operatorname{op}}
		\left\|
		g_{\infty,\lambda_{i+1}}
		-
		g_{\infty,\lambda_i}
		\right\|_\infty.
		\label{equation: uniform drift decomposition}
	\end{align}
	
	For the finite-width approximation term, the resolvent identity gives
	\begin{align*}
		g_{M,\lambda_i}-g_{\infty,\lambda_i}
		&=
		\lambda_i
		(L_\infty+\lambda_iI)^{-1}
		(L_M-L_\infty)
		(L_M+\lambda_iI)^{-1}
		g_\rho.
	\end{align*}
	Consequently,
	\begin{align*}
		\left\|
		g_{M,\lambda_i}-g_{\infty,\lambda_i}
		\right\|_\infty
		&\leq
		\lambda_i^{-1}
		\|L_M-L_\infty\|_{\operatorname{op}}
		\|g_\rho\|_\infty.
	\end{align*}
	Since
	\[
	\lambda_i^{-1}
	=
	\eta(i+1)^{1-\theta},
	\]
	Lemma~\ref{lemmaA} and the condition
	$M\geq T^{2(2-\theta)}$ imply that, with probability at least $1-\delta$,
	\begin{align}
		\sum_{i=1}^{t}
		\left\|
		g_{\infty,\lambda_i}-g_{M,\lambda_i}
		\right\|_\infty
		&\leq
		\|L_M-L_\infty\|_{\operatorname{op}}
		\|g_\rho\|_\infty
		\sum_{i=1}^{t}\lambda_i^{-1}
		\nonumber\\
		&\leq
		\kappa^2\eta2^{2-\theta}
		\|g_\rho\|_\infty
		\sqrt{2\log\frac{2}{\delta}}.
		\label{equation: finite width uniform bound}
	\end{align}
	
	For the regularization-path term, the reproducing property yields
	\[
	\|f\|_\infty
	\leq
	\kappa\|f\|_{\mathcal{H}_\infty}.
	\]
	Hence, by part~(2) of Lemma~\ref{chap7:lem:drift-lemma},
	\begin{align*}
		&
		\sum_{i=1}^{t-1}
		\left\|
		\prod_{j=i+1}^{t-1}
		\left(I-\eta_j(L_M+\lambda_jI)\right)
		\right\|_{\operatorname{op}}
		\left\|
		g_{\infty,\lambda_{i+1}}-g_{\infty,\lambda_i}
		\right\|_\infty
		\\
		&\leq
		\frac{
			2\kappa\|L_\infty^{-r}g_\rho\|_\rho
		}{
			2r-1
		}
		\sum_{i=1}^{t-1}
		\prod_{j=i+1}^{t-1}
		(1-\eta_j\lambda_j)
		\left|
		\lambda_{i+1}^{r-\frac12}
		-
		\lambda_i^{r-\frac12}
		\right|.
	\end{align*}
	Since
	\[
	\eta_j\lambda_j=\frac{1}{j+1},
	\]
	we have
	\[
	\prod_{j=i+1}^{t-1}
	(1-\eta_j\lambda_j)
	=
	\frac{i+1}{t}.
	\]
	Moreover, the mean value theorem gives
	\begin{align*}
		\left|
		\lambda_{i+1}^{r-\frac12}
		-
		\lambda_i^{r-\frac12}
		\right|
		&\leq
		\eta^{-r+\frac12}
		\left(r-\frac12\right)
		(1-\theta)
		(i+1)^{-(r-\frac12)(1-\theta)-1}.
	\end{align*}
	It follows that
	\begin{align*}
		&
		\sum_{i=1}^{t-1}
		\left\|
		\prod_{j=i+1}^{t-1}
		\left(I-\eta_j(L_M+\lambda_jI)\right)
		\right\|_{\operatorname{op}}
		\left\|
		g_{\infty,\lambda_{i+1}}-g_{\infty,\lambda_i}
		\right\|_\infty
		\\
		&\leq
		\frac{
			4\kappa\eta^{-r+\frac12}
			\|L_\infty^{-r}g_\rho\|_\rho
		}{
			(2r-1)
			\left(
			1-(r-\frac12)(1-\theta)
			\right)
		}
		t^{-(r-\frac12)(1-\theta)}.
	\end{align*}
	
	Combining this estimate with
	\eqref{equation: finite width uniform bound} and using
	$(a+b)^2\leq2a^2+2b^2$, we obtain
	\begin{align}
		&
		\left\|
		\sum_{i=1}^{t-1}
		\prod_{j=i+1}^{t-1}
		\left(I-\eta_j(L_M+\lambda_jI)\right)
		\left(g_{M,\lambda_{i+1}}-g_{M,\lambda_i}\right)
		\right\|_\infty^2
		\nonumber\\
		&\leq
		\Bigg[
		\frac{
			128\kappa^4\eta^2
		}{
			2^{2\theta}\theta^2
		}
		\|g_\rho\|_\infty^2
		+
		\frac{
			32\kappa^2\eta^{1-2r}
			\|L_\infty^{-r}g_\rho\|_\rho^2
		}{
			(2r-1)^2
			\left(2-(2r-1)(1-\theta)\right)^2
		}
		\Bigg]
		\log\frac{2}{\delta}.
		\label{chap7:eq:second-term-of-first-term-of-g_t-decomp}
	\end{align}
	
	It remains to estimate the stochastic term in
	\eqref{chap7:eq:temp-first-term-of-g_t-decomp}. By
	\eqref{eq:stochastic-term-induction} in the proof of
	Proposition~\ref{prop: uniform bound for E(g(t))}, and under the step-size
	condition \eqref{eq: conditon2 on eta}, we obtain
	\begin{align}
		\sum_{i=1}^{t-1}
		\eta_i^2
		\mathbb{E}_{Z^i}
		\left[
		\|\mathcal{B}_i\|_{\mathcal{H}_M}^2
		\right]
		&\leq
		\sum_{i=1}^{t-1}
		\eta_i^2
		\mathbb{E}_{Z^{i-1}}
		\left[
		\mathcal{E}(g^{(i)})
		\right]
		\nonumber\\
		&\leq
		\frac{2\eta^2\theta}{2\theta-1}
		\left(
		40\|g_\rho\|_\rho^2
		+
		2\mathcal{E}(g_\rho)
		\right)
		\nonumber\\
		&\leq
		\frac{1}{15\kappa^4}
		\left(
		20\|g_\rho\|_\rho^2
		+
		\mathcal{E}(g_\rho)
		\right).
		\label{chap7:eq:third-term-of-first-term-of-g_t-decomp}
	\end{align}
	
	Substituting
	\eqref{chap7:eq:second-term-of-first-term-of-g_t-decomp} and
	\eqref{chap7:eq:third-term-of-first-term-of-g_t-decomp} into
	\eqref{chap7:eq:temp-first-term-of-g_t-decomp} yields
	\begin{align*}
		\mathbb{E}_{Z^{t-1}}
		\left[
		\left\|g^{(t)}-g_{M,\lambda_t}\right\|_\infty^2
		\right]
		&\leq
		\Bigg[
		3\left(
		1+
		\frac{128\kappa^4\eta^2}
		{2^{2\theta}\theta^2}
		\right)
		\|g_\rho\|_\infty^2
		\\
		&\quad+
		\frac{
			96\kappa^2\eta^{1-2r}
			\|L_\infty^{-r}g_\rho\|_\rho^2
		}{
			(2r-1)^2
			\left(2-(2r-1)(1-\theta)\right)^2
		}
		\\
		&\quad+
		\frac{1}{5\kappa^2}
		\left(
		20\|g_\rho\|_\rho^2
		+
		\mathcal{E}(g_\rho)
		\right)
		\Bigg]
		\log\frac{2}{\delta}.
	\end{align*}
	Combining this estimate with \eqref{chap7:eq:g_t-decomp}, we obtain
	\begin{align}
		\mathbb{E}_{Z^{t-1}}
		\left[
		\left\|g^{(t)}\right\|_\infty^2
		\right]
		&\leq
		\Bigg[
		\left(
		8+
		\frac{256\kappa^4\eta^2}
		{2^{2\theta}\theta^2}
		\right)
		\|g_\rho\|_\infty^2+
		\frac{
			192\kappa^2\eta^{1-2r}
			\|L_\infty^{-r}g_\rho\|_\rho^2
		}{
			(2r-1)^2
			\left(2-(2r-1)(1-\theta)\right)^2
		}
		\nonumber\\
		&\quad+
		\frac{2}{5\kappa^2}
		\left(
		20\|g_\rho\|_\rho^2
		+
		\mathcal{E}(g_\rho)
		\right)
		\Bigg]
		\log\frac{2}{\delta}.
		\label{chap7:eq:final-g_t-decomp}
	\end{align}
	
	Finally, Assumption~\ref{assumption regularity condition} and
	$\frac12<r\leq1$ imply that
	\begin{align*}
		\|g_\rho\|_\infty
		\leq
		\kappa\|g_\rho\|_{\mathcal{H}_\infty}
		=
		\kappa
		\left\|
		L_\infty^{r-\frac12}
		L_\infty^{-r}g_\rho
		\right\|_\rho\leq
		\kappa^{2r}
		\left\|
		L_\infty^{-r}g_\rho
		\right\|_\rho.
	\end{align*}
	Substituting this bound into
	\eqref{chap7:eq:final-g_t-decomp} completes the proof.
\end{proof}

Now we are ready to prove the third term in (\ref{chap7:eq:error-decomp-of-convergence-error}).
\begin{proposition}
	\label{chap7:prop:third-term}
	Under the conditions of Theorem \ref{theorem general result}, when $M \geq T^{2(2-\theta)}$, the following inequality holds with probability at least $1-\delta$:
	\begin{equation}
		\label{chap7:eq:third-term}
		\sum_{t=1}^T \eta_t^2 \left( 1 +
		 \mathbb{E}_{Z^{t-1}} \left[\left\| g^{(t)} \right\|_{\infty}^2\right] \right) \operatorname{Tr} \left[ L_M^2 \prod_{i=t+1}^{T}\left(I-\eta_{i}(L_M + \lambda_i I)\right)^2 \right]
		\leq W_2 T^{-\theta} \log\frac{2}{\delta}, 
	\end{equation}
	where
	\begin{gather*}
		W_2 = \eta \kappa^2 \left[ 1 + \left(8 + \frac{256\kappa^4\eta^2 }{2^{2\theta}\theta^2} \right)\left\|g_{\rho} \right\|_{\infty}^2 + \frac{192\kappa^2 \eta^{1-2r} \left\| L_{\infty}^{-r} g_{\rho} \right\|_{\rho}^2}{(2r - 1)^2(2-(2r+1)(1-\theta))^2}  + \frac{2}{5\kappa^2 } \left( 20\| g_{\rho} \|_{\rho}^2 + \mathcal{E}\left(g_{\rho}\right) \right) \right]. 
	\end{gather*}
\end{proposition}
\begin{proof}
	First, we have 
	\begin{equation}\label{equ: error decomposition of the third term}
		\begin{aligned}
			&\sum_{t=1}^T \eta_t^2 \left( 1 + \mathbb{E}_{Z^{t-1}} \left[\left\| g^{(t)} \right\|_{\infty}^2\right] \right) \operatorname{Tr} \left[ L_M^2 \prod_{i=t+1}^{T}\left(I-\eta_{i}(L_M + \lambda_i I)\right)^2 \right]\\
			&\leq \left(1+\sup_{1\le t\le T}\left\{\mathbb{E}_{Z^{t-1}} \left[\left\| g^{(t)} \right\|_{\infty}^2\right] \right\}\right) \sum_{t=1}^T \eta_t^2  \operatorname{Tr} \left[ L_M^2 \prod_{i=t+1}^{T}\left(I-\eta_{i}(L_M + \lambda_i I)\right)^2 \right]\\
			&= \left(1+\sup_{1\le t\le T}\left\{\mathbb{E}_{Z^{t-1}} \left[\left\| g^{(t)} \right\|_{\infty}^2\right] \right\}\right) \operatorname{Tr} \left[ \sum_{t=1}^T \eta_t^2   L_M^2 \prod_{i=t+1}^{T}\left(I-\eta_{i}(L_M + \lambda_i I)\right)^2 \right]\\
			&\le \left(1+\sup_{1\le t\le T}\left\{\mathbb{E}_{Z^{t-1}} \left[\left\| g^{(t)} \right\|_{\infty}^2\right] \right\}\right)	\operatorname{Tr}(L_M) \left\| \sum_{t=1}^T \eta_t^2   L_M \prod_{i=t+1}^{T}\left(I-\eta_{i}(L_M + \lambda_i I)\right)^2 \right\|_{\operatorname{op}}. 
		\end{aligned}
	\end{equation}
		The last inequality follows from
	\[
	\operatorname{Tr}(AB)
	\leq
	\operatorname{Tr}(A)\|B\|_{\operatorname{op}}
	\]
	for every positive trace-class operator $A$ and every positive bounded
	operator $B$.
	
	It remains to estimate
	\[ \left\| \sum_{t=1}^T \eta_t^2   L_M \prod_{i=t+1}^{T}\left(I-\eta_{i}(L_M + \lambda_i I)\right)^2 \right\|_{\operatorname{op}}.\]
	Since all the operators involved are functions of the positive
	self-adjoint operator $L_M$, the spectral theorem yields	
	\begin{equation}
		\begin{aligned}
			&\left\| \sum_{t=1}^T \eta_t^2   L_M \prod_{i=t+1}^{T}\left(I-\eta_{i}(L_M + \lambda_i I)\right)^2 \right\|_{\operatorname{op}} \le \sup_{0<x\leq \kappa^2} \left\{ x \sum_{t=1}^T \eta_t^2 \prod_{i=t+1}^T (1-\eta_i(x + \lambda_i))^2 \right\}\\
			&\le  \sup_{0<x\leq \kappa^2} \left\{ \sup_{1\leq t \leq T} \left\{ \eta_t \prod_{i=t+1}^T (1-\eta_i(x + \lambda_i)) \right\} \times\sum_{t=1}^T \eta_t x \prod_{i=t+1}^T (1-\eta_i(x + \lambda_i)) \right\}\\
			&\le  \sup_{1\leq t \leq T} \left\{ \eta_t \prod_{i=t+1}^T (1-\eta_i \lambda_i) \right\} \times\sup_{0<x\leq \kappa^2} \left\{  \sum_{t=1}^T \eta_t x \prod_{i=t+1}^T (1-\eta_i(x + \lambda_i)) \right\}. 
		\end{aligned}
	\end{equation}	
		We estimate the two factors separately. Since
$
	\eta_t=\eta(t+1)^{-\theta},
	\lambda_t=\frac{1}{\eta}(t+1)^{-1+\theta},
$
	we have
$
	\eta_t\lambda_t=\frac{1}{t+1}.
$
	Therefore,
	\begin{align*}
		\sup_{1\leq t\leq T}
		\left\{
		\eta_t
		\prod_{i=t+1}^{T}(1-\eta_i\lambda_i)
		\right\}
		&=
		\sup_{1\leq t\leq T}
		\left\{
		\eta(t+1)^{-\theta}
		\prod_{i=t+1}^{T}\frac{i}{i+1}
		\right\}
		\\
		&=
		\sup_{1\leq t\leq T}
		\left\{
		\eta(t+1)^{-\theta}
		\frac{t+1}{T+1}
		\right\}
		\\
		&=
		\sup_{1\leq t\leq T}
		\left\{
		\frac{\eta(t+1)^{1-\theta}}{T+1}
		\right\}
		\\
		&\leq
		\eta(T+1)^{-\theta}
		\leq
		\eta T^{-\theta}.
	\end{align*}
	
	For the second factor, we use a telescoping argument:
	\begin{align*}
		\sum_{t=1}^{T}
		\eta_tx
		\prod_{i=t+1}^{T}
		(1-\eta_ix)=
		\sum_{t=1}^{T}
		\left[
		\prod_{i=t+1}^{T}(1-\eta_ix)
		-
		\prod_{i=t}^{T}(1-\eta_ix)
		\right]=
		1-\prod_{i=1}^{T}(1-\eta_ix)
		\leq1.
	\end{align*}
	Consequently,
	\begin{equation}
		\label{eq: middle term 1}
		\left\|
		\sum_{t=1}^{T}
		\eta_t^2L_M
		\prod_{i=t+1}^{T}
		\left(I-\eta_i(L_M+\lambda_iI)\right)^2
		\right\|_{\operatorname{op}}
		\leq
		\eta T^{-\theta}.
	\end{equation}
%
	Substituting the bound $\mathbb{E}_{Z^{t-1}} \left[\left\| g^{(t)} \right\|_{\infty}^2\right]$ proved in Proposition \ref{chap7:lem:uniform-boundedness-of-g_t}, together with
	\eqref{eq: middle term 1} and
	$\operatorname{Tr}(L_M)\leq\kappa^2$, into
	\eqref{equ: error decomposition of the third term} gives the desired result.
	\end{proof}

\subsection{Estimation of the Approximation Error}
For the approximation error, we have the following proposition.
\begin{proposition}
	\label{chap7:prop:approximation-error}
	Under the conditions of Theorem \ref{theorem general result}, when $M \geq T^{2(r+1)(1-\theta)}$, then with probability at least $1-\delta$,
	\begin{equation}
		\label{chap7:eq:approximation-error}
		\| g_{M, \lambda_{T+1}} - g_{\rho} \|_{\rho}^2 \leq 2\left(\eta^2 2^{3-2\theta} \kappa^4 \left\| g_{\rho} \right\|_{\rho}^2 +\eta^{-2r} \| L_\infty^{-r} g_{\rho} \|_{\rho}^2\right) T^{-2r(1-\theta)} \log \frac{2}{\delta}. 
	\end{equation}
\end{proposition}

\begin{proof}
		We begin with the decomposition
	\begin{equation}
		\label{chap7:eq:approximation-error-decomp}
		\| g_{M, \lambda_{T+1}} - g_{\rho} \|_{\rho}^2 \leq 2\| g_{M, \lambda_{T+1}} - g_{\infty, \lambda_{T+1}} \|_{\rho}^2 + 2\| g_{\infty, \lambda_{T+1}} - g_{\rho} \|_{\rho}^2. 
	\end{equation}
	We first bound the first term on the right-hand side of
	\eqref{chap7:eq:approximation-error-decomp}. By the definitions of
	$g_{M,\lambda_{T+1}}$ and $g_{\infty,\lambda_{T+1}}$ and the resolvent
	identity, we have
	\begin{align*}
		\left\| g_{M, \lambda_{T+1}} - g_{\infty, \lambda_{T+1}} \right\|_{\rho} &= \left\| \lambda_{T+1}(L_\infty + \lambda_{T+1} I)^{-1} (L_M-L_\infty)(L_M + \lambda_{T+1} I)^{-1} g_{\rho}\right\|_{\rho} \\
		&\le \lambda_{T+1}^{-1}\left\| L_M - L_\infty \right\|_{\operatorname{op}} \left\| g_{\rho} \right\|_{\rho},
	\end{align*}
		where we used
$
	\left\|(L_{\infty}+\lambda_{T+1}I)^{-1}\right\|_{\mathrm{op}}
	\leq \lambda_{T+1}^{-1}
$
	and
$
	\left\|(L_M+\lambda_{T+1}I)^{-1}\right\|_{\mathrm{op}}
	\leq \lambda_{T+1}^{-1}.
$
	
By part~(a) of Lemma~\ref{lemmaA}, with probability at least
$1-\delta$,
\[
\left\|L_M-L_{\infty}\right\|_{\mathrm{op}}
\leq
\kappa^{2}
\sqrt{\frac{2\log(2/\delta)}{M}}.
\]
Since
\[
\lambda_{T+1}
=
\frac{1}{\eta}(T+1)^{-(1-\theta)},
\]
it follows that
\begin{align*}
	\left\|g_{M,\lambda_{T+1}}
	-g_{\infty,\lambda_{T+1}}\right\|_{\rho} 
&\leq
	\eta(T+1)^{1-\theta}
	\kappa^{2}\|g_{\rho}\|_{\rho}
	\sqrt{\frac{2\log(2/\delta)}{M}} \\
	&\quad\leq
	\eta 2^{1-\theta}T^{1-\theta}
	\kappa^{2}\|g_{\rho}\|_{\rho}
	\sqrt{\frac{2\log(2/\delta)}{M}},
\end{align*}
where the last inequality follows from $T+1\leq 2T$ for $T\geq1$.

Therefore, if
\[
M\geq T^{2(r+1)(1-\theta)},
\]
then, with probability at least $1-\delta$,
\begin{equation}
	\label{chap7:eq:approximation-error-decomp-first-term}
	\begin{aligned}
		\left\|g_{M,\lambda_{T+1}}
		-g_{\infty,\lambda_{T+1}}\right\|_{\rho}^{2}
		\eta^{2}2^{3-2\theta}\kappa^{4}
		\|g_{\rho}\|_{\rho}^{2} 
		\times
		T^{-2r(1-\theta)}
		\log\frac{2}{\delta}.
	\end{aligned}
\end{equation}
	We next bound the second term on the right-hand side of
\eqref{chap7:eq:approximation-error-decomp}. By the definition of
$g_{\infty,\lambda_{T+1}}$, we have
\[
g_{\infty,\lambda_{T+1}}
=
(\lambda_{T+1}I+L_\infty)^{-1}L_\infty g_\rho.
\]
Therefore,
\begin{align*}
	g_{\infty,\lambda_{T+1}}-g_\rho
	&=
	\left(
	(\lambda_{T+1}I+L_\infty)^{-1}L_\infty-I
	\right)g_\rho \\
	&=
	-\lambda_{T+1}
	(\lambda_{T+1}I+L_\infty)^{-1}
	L_\infty^rL_\infty^{-r}g_\rho \\
	&=
	-\lambda_{T+1}
	(\lambda_{T+1}I+L_\infty)^{-1+r}
	(\lambda_{T+1}I+L_\infty)^{-r}
	L_\infty^rL_\infty^{-r}g_\rho.
\end{align*}
Since $1/2<r\le1$, Assumption~\ref{assumption regularity condition}
implies that $L_\infty^{-r}g_\rho\in\mathcal{L}^2_{\rho_X}$.
Moreover,
\[
\left\|
(\lambda_{T+1}I+L_\infty)^{-1+r}
\right\|_{\mathrm{op}}
\le
\lambda_{T+1}^{-(1-r)},
\]
and, by the spectral theorem,
\[
\left\|
(\lambda_{T+1}I+L_\infty)^{-r}
L_\infty^r
\right\|_{\mathrm{op}}
\le1.
\]
Hence,
\begin{align}
	\label{chap7:eq:approximation-error-decomp-second-term}
	\left\|
	g_{\infty,\lambda_{T+1}}
	-g_\rho
	\right\|_\rho^2
	\le
	\lambda_{T+1}^{2r}
	\left\|
	L_\infty^{-r}g_\rho
	\right\|_\rho^2 
	\le
	\eta^{-2r}
	(T+2)^{-2r(1-\theta)}
	\left\|
	L_\infty^{-r}g_\rho
	\right\|_\rho^2\le\eta^{-2r}
	T^{-2r(1-\theta)}
	\left\|
	L_\infty^{-r}g_\rho
	\right\|_\rho^2 .
\end{align}
Combining
\eqref{chap7:eq:approximation-error-decomp},
\eqref{chap7:eq:approximation-error-decomp-first-term}, and
\eqref{chap7:eq:approximation-error-decomp-second-term}
proves the desired result.
\end{proof}

\subsection{Proof of Theorem~\ref{theorem general result} and Corollary~\ref{corollary convergence rates}}
	We are now in a position to prove our main theorem.
\begin{proof}[Proof of Theorem~\ref{theorem general result}]
	Suppose that Assumptions~\ref{assumption activation function conditions},
	\ref{assumption uniform bounded}, and
	\ref{assumption regularity condition} (with $\frac12<r\le1$) hold.
	Throughout the proof, we assume that
	\[
	M\ge
	\max\left\{
	T^{6-5\theta},
	\,
	T^{2(2-\theta)},
	\,
	T^{2+2(r+1)(1-\theta)}
	\right\},
	\]
	so that all auxiliary results established in the previous subsections are applicable.
	
	Applying Lemma~\ref{chap7:lem:error-decomp-of-convergence-error}
	together with Propositions~\ref{chap7:prop:first-term},
	\ref{chap7:prop:second-term}, and
	\ref{chap7:prop:third-term}, we obtain
	\begin{equation}
		\label{chap7:eq:convergence-error}
		\begin{aligned}
			\mathbb{E}_{Z^{T}}
			\left[
			\left\|
			g^{(T+1)}-g_{\rho}
			\right\|_{\rho}^{2}
			\right]
			&\le
			3\|g_{\rho}\|_{\rho}^{2}T^{-\theta}
			+
			6V_1^{2}
			T^{-2r(1-\theta)}
			\log\frac{2}{\delta}
			+
			6V_2
			T^{-\theta}
			\log\frac{2}{\delta} \\
			&\le
			\left(
			3\|g_{\rho}\|_{\rho}^{2}
			+
			6V_1^{2}
			+
			6V_2
			\right)
			\left(
			T^{-2r(1-\theta)}
			+
			T^{-\theta}
			\right)
			\log\frac{2}{\delta}.
		\end{aligned}
	\end{equation}
	
	Next, applying the error decomposition
	\eqref{equation error decomposition}
	together with
	\eqref{chap7:eq:convergence-error},
	Proposition~\ref{proposition: dynamics error},
	and Proposition~\ref{chap7:prop:approximation-error},
	we obtain
	\[
	\mathbb{E}_{Z^{T}}
	\left[
	\left\|
	g_{\Theta^{(T+1)}}-g_{\rho}
	\right\|_{\rho}^{2}
	\right]
	\le
	\widetilde{C}
	\left(
	T^{-2r(1-\theta)}
	+
	T^{-\theta}
	\right)
	\log\frac{2}{\delta},
	\]
	where
	\[
	\widetilde{C}
	=
	147C_{\sigma}^{2}
	+
	3\left(
	3\|g_{\rho}\|_{\rho}^{2}
	+
	6V_1^{2}
	+
	6V_2
	\right)
	+
	6\left(
	2^{3-2\theta}\eta^{2}\kappa^{4}
	\|g_{\rho}\|_{\rho}^{2}
	+
	\eta^{-2r}
	\|L_{\infty}^{-r}g_{\rho}\|_{\rho}^{2}
	\right).
	\]
	This completes the proof.
\end{proof}

\begin{proof}[Proof of Corollary~\ref{corollary convergence rates}]
	Choosing
	\[
	\theta=\frac{2r}{2r+1}
	\]
	balances the two dominant terms,
	\[
	T^{-2r(1-\theta)}
	\quad\text{and}\quad
	T^{-\theta},
	\]
	since
	\[
	2r(1-\theta)=\theta=\frac{2r}{2r+1}.
	\]
	Therefore,
	\[
	T^{-2r(1-\theta)}
	=
	T^{-\theta}
	=
	T^{-\frac{2r}{2r+1}},
	\]
	which immediately yields the desired convergence rate.
\end{proof}

\newpage
\section{Appendix}\label{section: appendix}
\begin{appendices}
	In this section, we give some useful lemmas used in the proof of our main results.
	\begin{lemma}
		\label{lemmaA}
		For arbitrary $\delta \in (0, 1)$ and $M \in \mathbb{Z}_{+}$, the following inequality holds with confidence at least $1 - \delta$,
		\begin{enumerate}[$(a)$]
			\item $\left\| L_M - L_\infty \right\|_{op} \leq \| k_M - k_{\infty} \|_{L_{\infty}(\rho_\X \times \rho_\X)} \leq \kappa^2 \sqrt{\frac{2\log \frac{2}{\delta}}{M}}$. 
			\item $\left| \text{Tr} \left( L_M - L_\infty \right) \right| \leq \kappa^2 \sqrt{\frac{2\log \frac{2}{\delta}}{M}}$. 
			\item $\left\| L_M - L_\infty \right\|_{HS} \leq \kappa^2 \sqrt{\frac{2\log \frac{2}{\delta}}{M}}$. 
		\end{enumerate}
	\end{lemma}
	\begin{proof}
		\begin{enumerate}[$(a)$]
			\item Let's establish the proof for the first inequality. From the definition of operator norm, we get that 
			\begin{equation*}
				\begin{aligned}
					\left\| L_{M} - L_{\infty} \right\|_{op} &= \sup_{\| f \|_{\rho} = 1} \left\| L_{M} f - L_{\infty} f \right\|_{\rho} \\
					&= \sup_{\| f \|_{\rho} = 1} \left\| \int_{\mathcal{X}} f(x) (k_{M, x} - k_{\infty, x}) d\rho_{\X}(x) \right\|_{\rho} \\
					&= \sup_{\| f \|_{\rho} = 1} \sqrt{  \int_{\mathcal{X}} \left[ \int_{\mathcal{X}} f(x) \left(k_M(x, x^{\prime}) - k_{\infty}(x, x^{\prime})\right)d\rho_{\X}(x) \right]^2 d\rho_{\X}(x^{\prime}) } \\
					&\leq \sup_{\| f \|_{\rho} = 1} \sqrt{  \int_{\mathcal{X}} \left[ \left( \int_{\mathcal{X}} |f(x)|^2 dx\right) \left( \int_{\mathcal{X}} |k_M(x, x^{\prime}) - k_{\infty}(x, x^{\prime})|^2  d\rho_{\X}(x) \right)
						\right]  d\rho_{\X}(x^{\prime}) } \\
					&\leq \| k_M - k_{\infty} \|_{L_{\infty}(\rho_\X \times \rho_\X)}. 
				\end{aligned}
			\end{equation*}
			Thus we get the first half of the inequality.
			For the second half, for any $x, x^{\prime} \in \mc{X}$, define 
			\begin{equation*}
				\xi_r(x, x^{\prime}) = \sigma\left(b_r^{(0) \top} x\right) \sigma\left(b_r^{(0) \top} x^{\prime}\right) + \left(x^{\top} x^{\prime}+\gamma^{2}\right) \sigma^{\prime}\left(b_r^{(0) \top} x\right) \sigma^{\prime}\left(b_r^{(0) \top} x^{\prime}\right), 
			\end{equation*}
			then $\left| \xi_r(x, x^{\prime}) \right| \leq 4 + 2C_\sigma^2=\kappa^2$ holds for any $r = 1, \cdots, M$. Then, by applying the Hoeffding inequality (See e.g. Theorem 1 in \cite{boucheron2003concentration}), we can assert that
			\begin{equation*}
				| k_M(x, x^{\prime}) - k_{\infty}(x, x^{\prime}) | = \left| \frac{1}{M} \lsum_{r=1}^M \left(\xi_r(x, x^{\prime}) - \mb{E}_{b^{(0)}} \xi(x, x^{\prime})\right)\right| \leq \kappa^2 \sqrt{\frac{2\log \frac{2}{\delta}}{M}}. 
			\end{equation*}
			holds with confidence at least $1 - \delta$, 
			Thus $\| k_M - k_{\infty} \|_{L_{\infty}(\rho_\X \times \rho_\X)} \leq \kappa^2 \sqrt{\frac{2\log \frac{2}{\delta}}{M}}$. 
			\item This can be proved by utilizing the equation $\text{Tr}(L_\infty) = \int_{\mc{X}} k_{\infty}(x, x) d\rho_\X(x)$ and $(a)$. 
			\item According to the definition and properties of the norm of Hilbert-Schmidt operators, we can derive the following result
			\begin{equation*}
				\left\| L_M - L_\infty \right\|_{HS}^2 = \left| \text{Tr} \left[ (L_M - L_\infty)^* (L_M - L_\infty) \right] \right| \leq \left\| L_M - L_\infty \right\|_{op} \left| \text{Tr} \left( L_M - L_\infty \right) \right|. 
			\end{equation*}
			Therefore, the desired result follows directly from  $(a)$ and $(b)$.
		\end{enumerate}
	\end{proof}
	
	\begin{lemma}
		\label{lemmaB}
		Let $\eta_t = \eta t^{-\theta}$ with $\eta > 0$ and $\frac{1}{2} \le \theta < 1$, we have\\
		$(a)$ $\frac{\eta (1 - 2^{\theta - 1})}{1 - \theta} T^{1 - \theta} \leq \sum\limits_{t=1}^T \eta_t \leq \frac{\eta}{1 - \theta} T^{1 - \theta}$. \\
		$(b)$ $\sum\limits_{t=1}^T \eta_t^2 \leq \frac{2\eta^2 \theta}{  2\theta-1}$. \\
	\end{lemma}
	\begin{proof}
		The proof of $(a)$ and $(b)$ is standard, see e.g. Lemma 4.3 in \cite{guo2019fast}. 
	\end{proof}

\end{appendices}



\bibliographystyle{siam}
\bibliography{ref}

\end{document}